\documentclass[letterpaper]{article} 
\usepackage{aaai24}  
\usepackage{times}  
\usepackage{helvet}  
\usepackage{courier}  
\usepackage[hyphens]{url}  
\usepackage{graphicx} 
\usepackage{natbib}  
\usepackage{caption} 
\usepackage{algorithm}
\usepackage{algorithmic}

\usepackage{newfloat}
\usepackage{listings}
\DeclareCaptionStyle{ruled}{labelfont=normalfont,labelsep=colon,strut=off} 
\floatstyle{ruled}
\newfloat{listing}{tb}{lst}{}
\floatname{listing}{Listing}
\usepackage{booktabs}
\usepackage{epsfig}
\usepackage{amsmath}
\usepackage{amssymb}
\usepackage{pifont}
\usepackage{multirow}
\usepackage{multicol}
\usepackage{xcolor}
\usepackage{xspace}
\usepackage{enumitem}
\usepackage{color, colortbl}
\usepackage{pdfpages}
\usepackage{enumitem}
\usepackage{subfiles}

\setlist{nosep}
\usepackage[capitalize]{cleveref}
\crefname{section}{Sec.}{Secs.}
\Crefname{section}{Section}{Sections}
\Crefname{table}{Table}{Tables}
\crefname{table}{Tab.}{Tabs.}

\definecolor{lightgray}{gray}{0.9}

\makeatletter
\DeclareRobustCommand\onedot{\futurelet\@let@token\@onedot}
\def\@onedot{\ifx\@let@token.\else.\null\fi\xspace}
\def\eg{\emph{e.g}\onedot} 
\def\ie{\emph{i.e}\onedot}

\def\etal{\emph{et al}\onedot}
\makeatother

\title{MaskDiff: Modeling Mask Distribution with Diffusion Probabilistic Model \\ for Few-Shot Instance Segmentation}
\author{Minh-Quan Le\textsuperscript{\rm 1, \rm 2, \rm 3}, Tam V. Nguyen\textsuperscript{\rm 4}, Trung-Nghia Le\textsuperscript{\rm 1, \rm 2}, \\Thanh-Toan Do\textsuperscript{\rm 5},
Minh N. Do\textsuperscript{\rm 6}, Minh-Triet Tran\textsuperscript{\rm 1, \rm 2}\thanks{Corresponding author}\\
}
\affiliations {
    \textsuperscript{\rm 1} University of Science, VNU-HCM, Ho Chi Minh City, Vietnam \\
    \textsuperscript{\rm 2} Vietnam National University, Ho Chi Minh City, Vietnam\\
    \textsuperscript{\rm 3} Stony Brook University, United States\\
    \textsuperscript{\rm 4} University of Dayton, United States\\
    \textsuperscript{\rm 5} Monash University, Australia\\
    \textsuperscript{\rm 6} University of Illinois at Urbana-Champaign, United States\\
    lmquan@selab.hcmus.edu.vn, tamnguyen@udayton.edu, ltnghia@fit.hcmus.edu.vn, \\toan.do@monash.edu, minhdo@illinois.edu, tmtriet@fit.hcmus.edu.vn
}

\begin{document}

\maketitle

\begin{abstract}
Few-shot instance segmentation extends the few-shot learning paradigm to the instance segmentation task, which tries to segment instance objects from a query image with a few annotated examples of novel categories. Conventional approaches have attempted to address the task via prototype learning, known as point estimation. However, this mechanism depends on prototypes (\eg mean of $K-$shot) for prediction, leading to performance instability. To overcome the disadvantage of the point estimation mechanism, we propose a novel approach, dubbed MaskDiff, which models the underlying conditional distribution of a binary mask, which is conditioned on an object region and $K-$shot information. Inspired by augmentation approaches that perturb data with Gaussian noise for populating low data density regions, we model the mask distribution with a diffusion probabilistic model. We also propose to utilize classifier-free guided mask sampling to integrate category information into the binary mask generation process. Without bells and whistles, our proposed method consistently outperforms state-of-the-art methods on both base and novel classes of the COCO dataset while simultaneously being more stable than existing methods. The source code is available at: https://github.com/minhquanlecs/MaskDiff.
\end{abstract}

\section{Introduction}

\begin{figure}[t!]
    \centering
    \includegraphics[width=\linewidth]{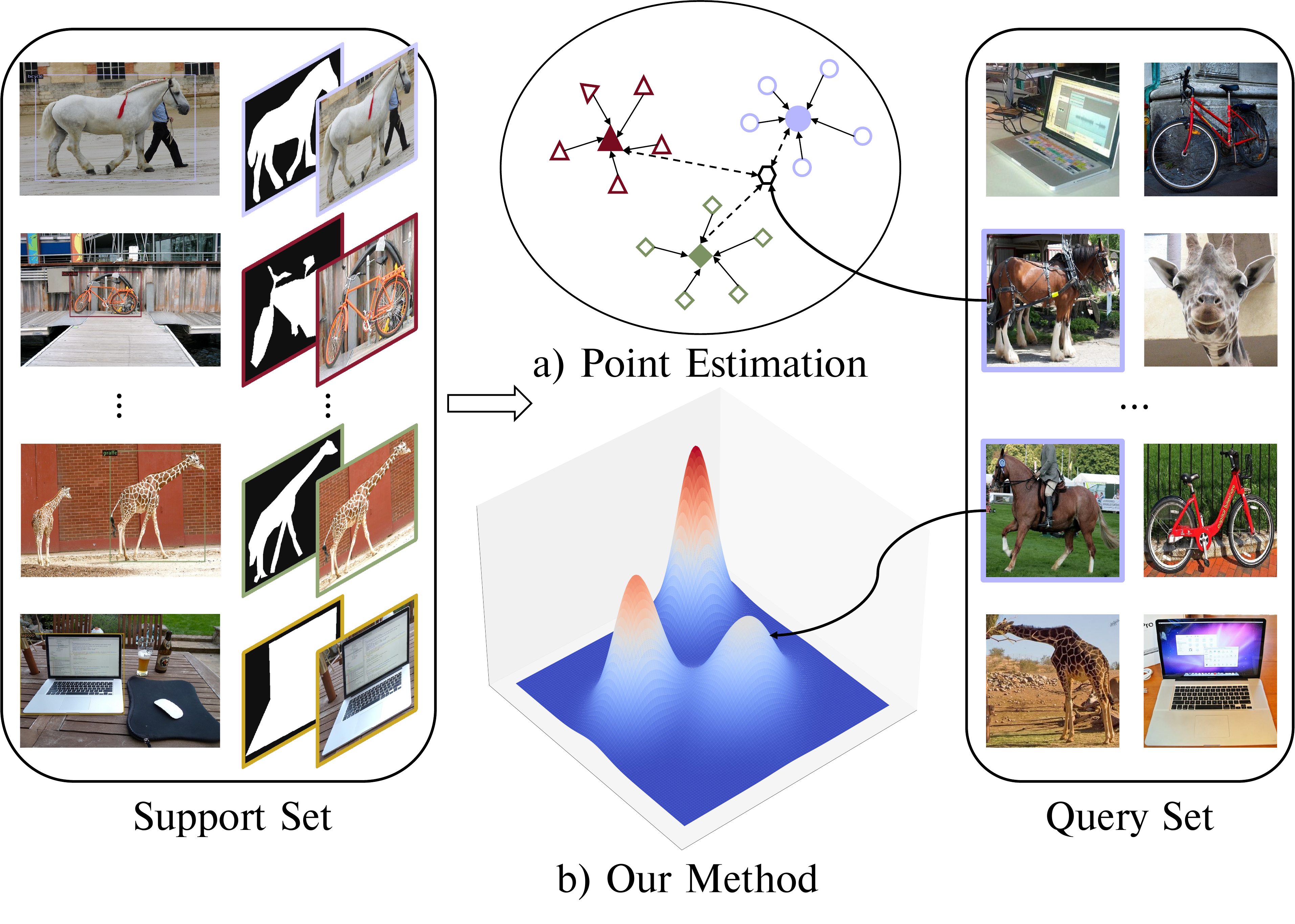}
     \vspace{-2mm}
    \caption{(a) Prior work estimates a specific point as representative of each class~\cite{Fan-CVPR2020}. (b) Our MaskDiff method models the conditional distribution of masks given object regions and $K-$shot samples.}
    \label{fig:distribution}
    \vspace{-5mm}
\end{figure}

To achieve outstanding performances, instance segmentation models~\cite{He-ICCV2017,ltnghia-TIP2022} require to be trained on substantial pixel-level annotated images, which is labor-intensive and costly. Moreover, their ability to generalize from a few examples is still far from acceptable compared to the human visual system, which constrains their feasibility in practical applications. Inspired by the miraculous ability of the human visual system to recognize novel objects with limited data, few-shot learning (FSL) aims to learn new concepts on a handful of training data ($K$ examples) by generalizing models trained on base classes to adapt novel classes. 

Conventional few-shot instance segmentation (FSIS) methods tried to address learning on a few examples via prototype learning~\cite{Fan-CVPR2020, Ganea-CVPR2021}. This matching mechanism searches the nearest prototype and its corresponding support class from support images as guidance for segmenting the query image (Fig.\ref{fig:distribution}a). Unlike existing FSIS methods, we introduce a novel probabilistic model to estimate the distribution of a binary mask given a detected object region and $K$ samples (Fig.\ref{fig:distribution}b). Concretely, object region, object category, and $K-$shot are treated as conditional information for generating the binary mask representation of each object. Mathematically, we model the conditional distribution of a binary mask of an object instance conditioned on an object region and $K-$shot of a specific $i$-th class  $p_\theta\left(\mathbf{y}_{\mathrm{mask}}\vert \mathbf{x}_{\mathrm{region}}, K\mathrm{-shot}^i\right)$. To the best of our knowledge, our work is the first method to model mask distribution for FSIS.

Derived from the insight that perturbs data with Gaussian noise for populating low data density regions~\cite{Song-NIPS2019} and motivated by diffusion-based models~\cite{Dickstein-ICML2015}, we define a Markov chain of diffusion steps to slowly add Gaussian noise to an object mask and then learn to reverse the diffusion process conditioned on an object region and $K-$shot samples to reconstruct the desired mask representation corresponding to the objects from the noise. We also design a method using classifier-free guidance to guide the diffusion model during sampling. Hence, the diffusion model is aware of categories of binary masks in the sampling procedure.

Our proposed approach includes several advantages. First, our method is more stable and robust to different $K-$shot samples than point estimation methods since we capture the underlying conditional distribution of mask representation rather than depending on prototypes for prediction (see subsection stability analysis). We hypothesize that the binary mask representation of each object is sampled from a high-dimensional conditional distribution conditioned on RGB image and $K-$shot. Thus, a single prototypical vector is insufficient to comprehensively capture diverse levels of semantic information, including object boundaries, poses, or categories. In addition, we argue that the mask representation relies on not only the instance object region but also its category. Therefore, we follow class-specific mask predictor and integrate category knowledge into the mask generation process.
Last but not least, our class-specific mask predictor alleviates spatial information losses from pooling operators of existing FSIS methods~\cite{Ganea-CVPR2021, Nguyen-CVPR2022}. When exploiting detected bounding box locations as input for diffusion models, we do not leverage any pooled features such as Mask-RCNN~\cite{He-ICCV2017} to generate masks but directly use image channels, thus preserving detailed spatial information. Our contributions lie in four-fold:
\begin{itemize}
\item We introduce a novel mask distribution modeling approach, dubbed \textbf{MaskDiff}, for FSIS. Conceptually, unlike point estimation, MaskDiff tries to model the conditional distribution of a binary mask conditioned on the detected object region and $K-$shot samples. 
    \item Our work is the first to adapt conditional diffusion probabilistic models for modeling instance binary mask distribution in FSIS.
    \item We propose to utilize guidance from the available classification head of the object detector to integrate category information into the mask sampling procedure. Consequently, in the reverse process, the diffusion model is aware of the categories of mask representation. Our class-specific mask predictor outperforms class-agnostic ones~\cite{Ganea-CVPR2021, Nguyen-CVPR2022}. Moreover, when we integrate category information into the mask-generation process, the sampling procedure becomes more stable and produces more organized content (see Supplementary Materials for qualitative results).
    \item Extensive experimental results on the standard COCO dataset show that our MaskDiff not only outperforms state-of-the-art FSIS methods on both base and novel classes but also is more stable than existing methods. 
\end{itemize}

\section{Related Work}
\textbf{Few-shot instance segmentation (FSIS).} The majority of prior works for FSIS try to provide guidance to specific components of Mask-RCNN~\cite{He-ICCV2017} to guarantee that the networks better understand novel classes or both base and novel classes as well. FGN~\cite{Fan-CVPR2020} integrates multiple guiding techniques into the core components of Mask R-CNN, such as the masked pooling operator applied before computing the class-attentive vectors and generating masks. Most early approaches adapted meta-learning to episodic training~\cite{Michaelis-2018,Yan-ICCV2019, Fan-CVPR2020}. Following this training direction, few modern methods utilized anchor-free object detectors to directly segment objects from arbitrary categories given by reference images~\cite{Nguyen-CVPR2021, Wang-MM2022}. On the other hand, Wang \etal~\shortcite{Wang-ICML2020} proposed a two-stage fine-tuning approach (TFA) which first trains Faster-RCNN~\cite{Shaoqing-NIPS2015} on the base classes and then fine-tunes the box predictor on a balanced subset of base and novel classes. Modern FSIS methods~\cite{Wang-ICML2020,Ganea-CVPR2021,Nguyen-CVPR2022} were then developed in this fine-tuning direction by fine-tuning the last layers of certain heads on novel classes. Our proposed MaskDiff also follows the latter training strategy. In addition, current techniques~\cite{Ganea-CVPR2021,Nguyen-CVPR2022} mainly concentrate on class-agnostic mask prediction heads, meaning each object's mask representation does not depend on its category. On the contrary, we propose to leverage object classes to generate semantic and stable masks.  

\textbf{Diffusion probabilistic model (DPM).} 
Diffusion models~\cite{Dickstein-ICML2015} belong to a class of likelihood-based generative models that comprise two processes, namely, the forward diffusion process and the reverse diffusion process. 
Ho \etal~\shortcite{Ho-NIPS2020} first clarified the equivalence of diffusion models and score-based models~\cite{Song-NIPS2019} and proposed a denoising DPM. Later on, further techniques~\cite{
Dhariwal-NIPS2021} were designed to improve and illustrate the superior performance of diffusion models over GAN~\cite{Goodfellow-NIPS2014}. In particular, an additional classifier is trained on noisy images and utilizes gradients to guide the sampling process based on the conditioning information. On the contrary, Ho and Salimans~\shortcite{Ho-NIPSW2021} argued that a pure generative model could provide guidance without needing a classifier. Inspired by the work of Ho and Salimans~\shortcite{Ho-NIPSW2021}, we employ classifier-free guided diffusion-based mask sampling, using the off-the-shelf cosine classifier from the box-classification head instead of training additional classifiers.


Several research works have studied the adaptation of conditional DPMs to downstream tasks including super-resolution~\cite{Ho-JMLR2022,Rombach-CVPR2022}, text-to-image synthesis~\cite{Kim-CVPR2022,Gu-CVPR2022}, image inpainting~\cite{Lugmayr-CVPR2022, Chung-CVPR2022}, and image segmentation~\cite{Baranchuk-ICLR2022}. To the best of our knowledge, our proposed MaskDiff is the first work that adapts conditional DPM to FSIS.

\section{Methodology}
\subsection{Problem Formulation}
In few-shot learning~\cite{Kang-ICCV2019}, we have a disjoint set of base classes $\mathbf{C}_{\mathrm{base}}$ which contains a large quantity of training data and a set of novel classes $\mathbf{C}_{\mathrm{novel}}$ with a limited number of annotated data $\mathbf{C}_{\mathrm{base}} \cap \mathbf{C}_{\mathrm{novel}} = \emptyset$. The main objective is to train a model that performs well on the novel classes $\mathbf{C}_{\mathrm{test}} = \mathbf{C}_{\mathrm{novel}}$ or on both base and novel classes together $\mathbf{C}_{\mathrm{test}} = \mathbf{C}_{\mathrm{base}} \cup \mathbf{C}_{\mathrm{novel}}$. Following the steps of few-shot classification, prior works~\cite{Kang-ICCV2019,Yan-ICCV2019} simulate the \textit{episodic-training} methodology which randomly samples a series of episodes $\mathcal{E} = \left\{\left(\mathbf{I}_j^q, \mathcal{S}_j\right)\right\}_{j=1}^{\left|\mathcal{E}\right|}$ where the $j$-th episode is formulated by sampling a support set $\mathcal{S}_j$ including $N$ classes from $\mathbf{C}_{\mathrm{train}} = \mathbf{C}_{\mathrm{base}} \cup \mathbf{C}_{\mathrm{novel}}$ in tandem with $K$ samples per class ($N-$way, $K-$shot) and a query image $\mathbf{I}^q_j$. The goal of FSIS is not only to localize all the object instances from any query image and classify those objects but also to determine their segmentation masks. For all objects in a query image $\mathbf{I}^q$ that belong to $\mathbf{C}_{\mathrm{test}}$, FSIS generates the corresponding labels, bounding boxes, and segmentation masks. 
\begin{figure}[!t]
    \centering
    \includegraphics[width=\linewidth]{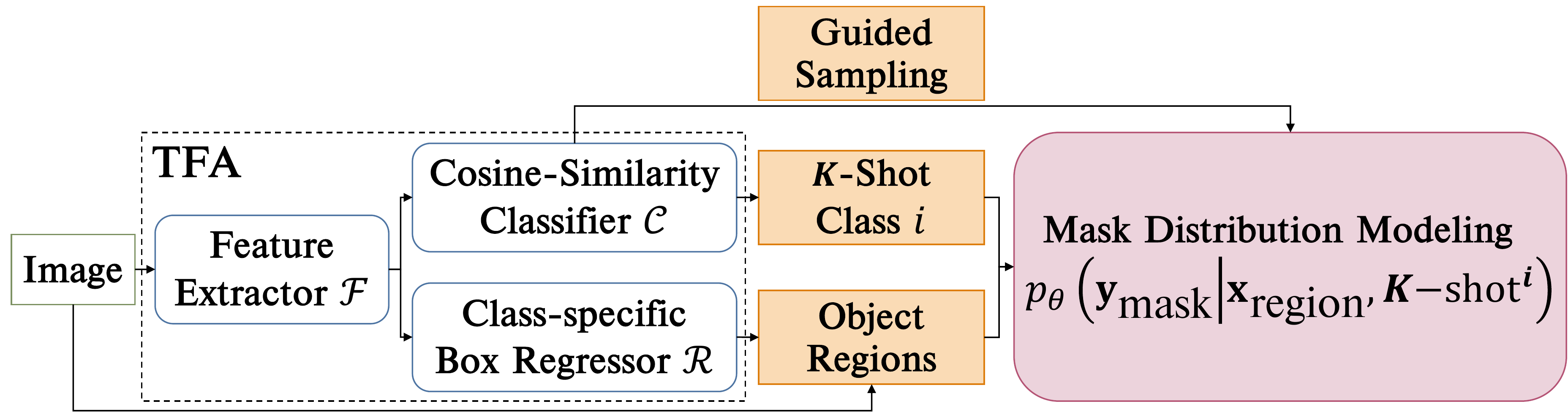}
    \caption{Our MaskDiff is built upon TFA~\cite{Wang-ICML2020} by integrating a mask distribution modeling head and adapting guided sampling to integrate category information into mask generation procedure.}
    \label{fig:architecture}
    \vspace{-6mm}
\end{figure}

\subsection{Proposed MaskDiff Method}
\textbf{Overview.} Figure \ref{fig:architecture} illustrates the architecture of our MaskDiff, adapted from TFA~\cite{Wang-ICML2020}, which is a two-stage object detection architecture. We build a mask distribution modeling head on top of TFA. Unlike class-agnostic mask predictor~\cite{Ganea-CVPR2021} or prototype-based mask head~\cite{Fan-CVPR2020}, we propose to model the conditional distribution of a binary mask conditioned on an instance region and $K-$shot samples via conditional DPMs. With regard to denoising architecture of the mask distribution modeling branch, we adapt UNet~\cite{Ronneberger-MICCAI2015} architecture with attention heads~\cite{Vaswani-NIPS2017} from Guided Diffusion~\cite{Dhariwal-NIPS2021}. In particular, the conditioning module consists of concatenating a detected object region and $K-$shot. We empirically found that the permutation of $K-$shot does not affect the model's performance since the encoder-decoder architecture leverages information of each shot independently. The detailed denoising architecture is illustrated in Supplementary Material (Supp.).

\begin{figure}[!t]
    \centering
    \includegraphics[width=\linewidth]{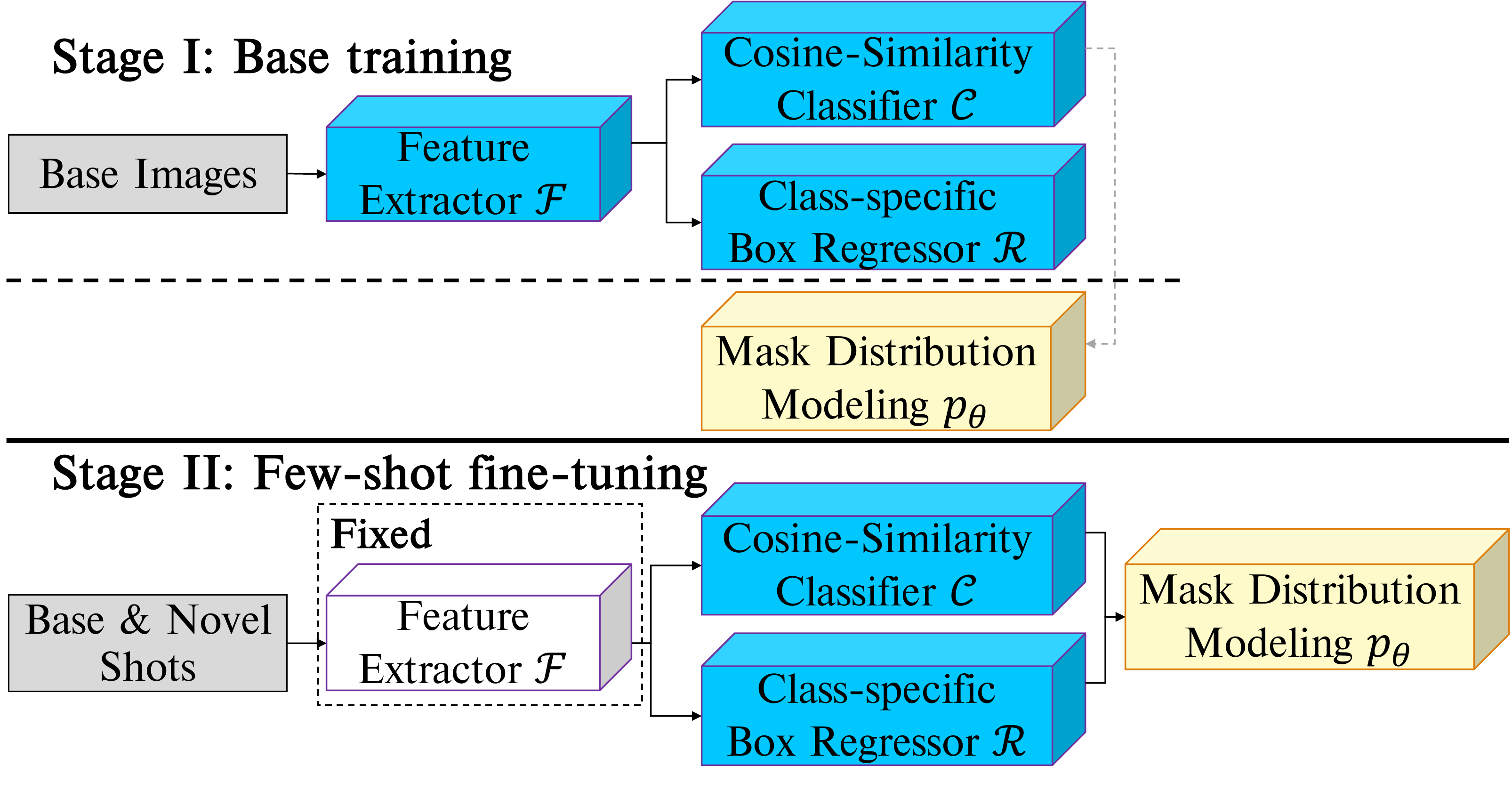}
    \vspace{-5mm}
    \caption{Pipeline of diffusion-based two-stage training. In the first stage, the entire object detector (feature extractor, classifier, and box regressor) and the mask distribution modeling head are trained separately on the base classes. In the second stage, the feature extractor is frozen while the cosine classifier, the box regressor, and the mask distribution modeling head are jointly fine-tuned on both base and novel classes. Our mask distribution modeling head is trained based on diffusion (yellow), while the remaining two heads follow the standard object detection training (blue).}
    \label{fig:approach}
    \vspace{-3mm}
\end{figure}

\textbf{Diffusion-based two-stage training.} Figure~\ref{fig:approach} visualizes our diffusion-based two-stage training for FSIS. In the first stage, \ie, the base training stage, the network is trained only on the base classes $\mathbf{C}_{\mathrm{base}}$. We separately train few-shot object detector heads and estimate the mask distribution. Specifically, the RoI cosine-similarity classifier $\mathcal{C}$ and the box regressor $\mathcal{R}$ follow the standard training process while the mask distribution modeling head $p_\theta\left(\mathbf{y}_{\mathrm{mask}}\vert \mathbf{x}_{\mathrm{region}}, K\mathrm{-shot}^i\right)$ is proposed to train based on conditional DPM. The reason is that in this early stage, the box regressor $\mathcal{R}$ is not stable enough, and it is complicated for probabilistic models to estimate the mask distribution efficiently. Obviously, incorrectly localizing objects leads to extremely unsatisfactory segmentation results. 

The second stage, \ie, the few-shot fine-tuning stage, involves freezing the entire feature extractor $\mathcal{F}$ and jointly fine-tuning prediction heads on a balanced dataset of $K$ shots for both $\mathbf{C}_{\mathrm{base}}$ and $\mathbf{C}_{\mathrm{novel}}$ classes. Likewise, all the prediction heads $\mathcal{C}$, $\mathcal{R}$, and $p_\theta$ are fine-tuned. 

\subsection{Instance Mask Modeling with Conditional DPM}
\begin{figure}[!t]
    \centering
    \includegraphics[width=\linewidth]{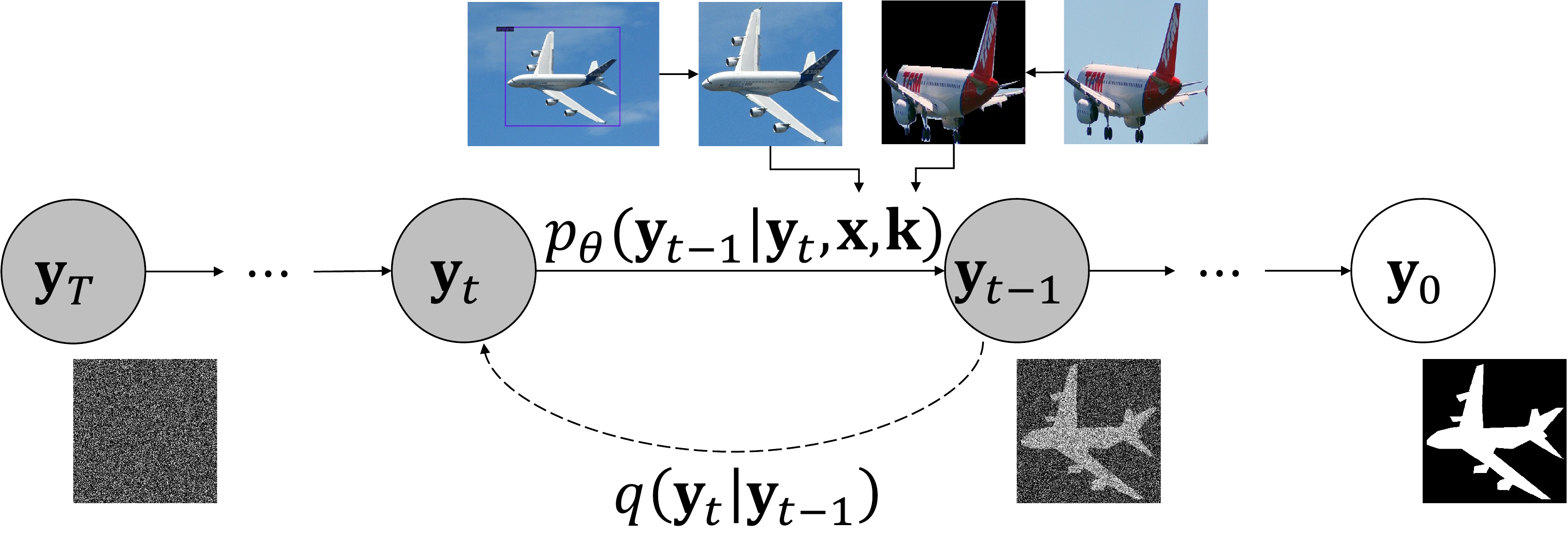}
    \vspace{-5mm}
    \caption{Conditional diffusion model for few-shot binary mask segmentation. The directed graphical model transforms the noise from standard Gaussian distribution to the mask representation of objects through an iterative denoising process. Solid arrows indicate the reverse process, while the dashed arrow implies the forward step. $\mathbf{y},\mathbf{x},\mathbf{k}$ denote binary masks, object regions and removed background version of $K-$shot samples, respectively.}
    \label{fig:diffusion}
    \vspace{-4mm}
\end{figure}
Given an FSIS dataset, we extract a dataset of input-output pairs denoted $\mathcal{D} = \left\{\left(\mathbf{x}_{\mathrm{region}}^d, K\mathrm{-shot}; \mathbf{y}_{\mathrm{mask}}^d\right)\right\}_{d=1}^{\left|\mathcal{D}\right|}$ for modeling the binary mask distribution given object regions and $K-$shot information. $\mathbf{x}^d_{\mathrm{region}}$'s are collected by cropping object regions from the RGB images by bounding box locations in the annotations. $\mathbf{y}^d_{\mathrm{mask}}$'s are converted from polygons to binary mask representations at the same location as $\mathbf{x}^d_{\mathrm{region}}$'s. Finally, $K-$shot samples of each class are the background removed version of the object region. For simplicity, we denote $\mathbf{y}, \mathbf{x}, \mathbf{k}$ to represent binary masks, object regions, and $K-$shot guidance, respectively. In this work, we aim to learn the underlying conditional distribution from which a data point representing a binary mask is sampled $\mathbf{y}_0 \sim q(\mathbf{y}\vert \mathbf{x}, \mathbf{k})$. However, we are unable to determine exactly the true distribution. Via diffusion models, we maximize the likelihood $p_{\theta}\left(\mathbf{y}_{0}\vert \mathbf{x}, \mathbf{k}\right)=\int p_{\theta}\left(\mathbf{y}_{0: T}\vert \mathbf{x}, \mathbf{k}\right) d \mathbf{y}_{1: T}$ to approximate the true distribution. Two processes are defined in the conditional diffusion probabilistic models, including the forward and reverse processes (see Fig.~\ref{fig:diffusion}).

\begin{algorithm}[!t]
\caption{Training Procedure}\label{alg:training}
\small
\begin{algorithmic}
\REPEAT
\STATE $\mathbf{x}\sim q(\mathbf{x})$, $\mathbf{x}$ belongs to class $i$,  $\mathbf{k} = $ $K$ shots of class $i$ 
\STATE $\mathbf{y}_0\sim q(\mathbf{y}_0|\mathbf{x}, \mathbf{k})$
\STATE $t\sim \mathrm{Uniform}({1,\dots,T})$
\STATE $\boldsymbol{\epsilon}\sim \mathcal{N}(\mathbf{0}, \mathbf{I})$
\STATE Take gradient descent step on 
\begin{equation}
\nabla_{\theta}\left\|\boldsymbol{\epsilon}-\boldsymbol{\epsilon}_{\theta}\left(\sqrt{\bar{\alpha}_{t}} \mathbf{y}_{0}+\sqrt{1-\bar{\alpha}_{t}} \boldsymbol{\epsilon}, \mathbf{x}, \mathbf{k}, t\right)\right\|^{2} \notag
\end{equation}
\UNTIL{converged}
\end{algorithmic}
\end{algorithm}

\textbf{Forward diffusion process.} The forward diffusion process is defined in which we gradually add a small amount of Gaussian noise to the sample in $T$ steps, producing a sequence of noisy samples $\mathbf{y}_1, \dots, \mathbf{y}_T$, formulated as follows:
\begin{equation}
\footnotesize
    q\left(\mathbf{y}_{1: T} \vert \mathbf{y}_{0}\right)=\prod_{t=1}^{T} q\left(\mathbf{y}_{t} \vert \mathbf{y}_{t-1}\right).
\end{equation}

\textbf{Reverse diffusion process.} The reverse diffusion process $p_{\theta}\left(\mathbf{y}_{0: T} \vert \mathbf{x}, \mathbf{k}\right)$ is defined as a Markov chain with learned Gaussian transitions beginning with $p(\mathbf{y}_T) \sim \mathcal{N}(\mathbf{0}, \mathbf{I})$, which is formulated as follows:
\begin{equation}
\footnotesize
p_{\theta}\left(\mathbf{y}_{0:T} \vert \mathbf{x}, \mathbf{k}\right)=p\left(\mathbf{y}_{T}\right) \prod_{t=1}^{T} p_{\theta}\left(\mathbf{y}_{t-1} \vert \mathbf{y}_{t}, \mathbf{x}, \mathbf{k}\right).
\end{equation}

\textbf{Diffusion loss.} The conditional diffusion probabilistic model is trained to minimize the cross entropy as the learning objective, which is equivalent to minimizing variational upper bound $L_\mathrm{VUB}$ (see Supplementary Materials for detailed derivation of loss functions). 
The loss $L_\mathrm{VUB}$ can be rewritten to be a combination of several KL-divergence and entropy terms:
{\footnotesize	
\begin{align}
L_\text{VUB} &= \mathbb{E}_q \Bigg[\underbrace{D_\text{KL}(q(\mathbf{y}_T \vert \mathbf{y}_0) \parallel p_\theta(\mathbf{y}_T))}_{L_T} \underbrace{-\log p_\theta(\mathbf{y}_0 \vert \mathbf{y}_1, \mathbf{x}, \mathbf{k})}_{L_0} \notag\\
&+ \sum_{t=1}^{T-1} \underbrace{D_\text{KL}(q(\mathbf{y}_t \vert \mathbf{y}_{t+1}, \mathbf{y}_0) \parallel p_\theta(\mathbf{y}_t \vert\mathbf{y}_{t+1}, \mathbf{x}, \mathbf{k}))}_{L_{t}}\Bigg].
\end{align}}

Following the standard training process of DPM~\cite{Ho-NIPS2020}, the loss term $L_t$ is parameterized to achieve better training, resulting in a simplified loss:
\begin{equation}
\footnotesize
L_t = \mathbb{E}_{\mathbf{y}_0, \boldsymbol{\epsilon}} \Big[\|\boldsymbol{\epsilon} - \boldsymbol{\epsilon}_\theta(\sqrt{\bar{\alpha}_t}\mathbf{y}_0 + \sqrt{1 - \bar{\alpha}_t}\boldsymbol{\epsilon}, \mathbf{x}, \mathbf{k}, t)\|^2 \Big],
\end{equation}
where $\boldsymbol{\epsilon} \sim \mathcal{N}(\mathbf{0}, \mathbf{I})$. Training procedure is shown in Alg.~\ref{alg:training}.

\textbf{Classifier-free guided mask sampling.} In contrast to class-agnostic mask segmentation~\cite{Ganea-CVPR2021,Nguyen-CVPR2022}, we integrate category knowledge into binary mask generation. It is known that the mask representation of each object depends upon not only its boundary but also its category. Hence, the mask distribution modeling head can be aware of the discriminative properties of objects. In other words, the diffusion model explicitly controls binary masks via object labels. 

Inspired by the derivation of score-based models~\cite{Song-NIPS2019}, the distribution $p(\mathbf{y}_t|\mathbf{x}, \mathbf{k}, \mathbf{c})$, where $\mathbf{c}$ is a one-hot vector indicating the object category, has the score:
\begin{equation}
\footnotesize
\resizebox{0.9\hsize}{!}{$
\begin{aligned}
\nabla \log p\left(\mathbf{y}_t|\mathbf{x},\mathbf{k}, \mathbf{c}\right) &=\nabla \log \left(\frac{p\left(\mathbf{y}_t| \mathbf{x},\mathbf{k}\right) p\left(\mathbf{c} | \mathbf{y}_t,  \mathbf{x},\mathbf{k}\right)}{p(\mathbf{c} | \mathbf{x},\mathbf{k})}\right) \\
&=\underbrace{\nabla \log p\left(\mathbf{y}_t| \mathbf{x},\mathbf{k}\right)}_{\text {guidance-agnostic score }}+\underbrace{\nabla \log p\left(\mathbf{c} | \mathbf{y}_t, \mathbf{x},\mathbf{k}\right)}_{\text {adversarial gradient}}.
\end{aligned}$} \label{eq:cls_guide}
\end{equation}

\begin{algorithm}[!t]
\caption{Inference Procedure with Guided Sampling}\label{alg:inference}
\small
\begin{algorithmic}
\STATE $\mathbf{y}_T \sim \mathcal{N}(\mathbf{0}, \mathbf{I})$
\FOR{$t = T,\dots,1$} 
\STATE $\boldsymbol{\epsilon} \sim \mathcal{N}(\mathbf{0}, \mathbf{I})$ if $t > 1$, otherwise $\boldsymbol{\epsilon} \leftarrow \mathbf{0}$
\STATE $\boldsymbol{\hat{\epsilon}}_\theta \leftarrow \omega\boldsymbol{\epsilon}_\theta(\mathbf{y}_t, \mathbf{x}, \mathbf{k}, \mathbf{c}, t)+ (1-\omega)\boldsymbol{\epsilon}_\theta(\mathbf{y}_t, \mathbf{x}, \mathbf{k}, t)$ 
\STATE $\mathbf{y}_{t-1} \leftarrow \frac{1}{\sqrt{\alpha_t}} \Big( \mathbf{y}_t - \frac{\beta_t}{\sqrt{1 - \bar{\alpha}_t}} \boldsymbol{\hat{\epsilon}}_\theta \Big) + \sigma_t \boldsymbol{\epsilon}$
\ENDFOR
\RETURN $\mathbf{y}_0$
\end{algorithmic}
\end{algorithm}

\begin{table*}[t!]
\centering
\vspace{-2mm}
\resizebox{\linewidth}{!}{
\begin{tabular}{c|l|c|cc|cc|cc|cc|cc|cc}
\toprule
\multirow{3}{*}{\textbf{Shots}} &  \multirow{3}{*}{\textbf{Method}} & \multirow{3}{*}{\textbf{Published}} & \multicolumn{6}{c|}{\textbf{Object Detection}} & \multicolumn{6}{c}{\textbf{Instance Segmentation}} \\ 
 & & & \multicolumn{2}{c|}{\textbf{All Classes}} & \multicolumn{2}{c|}{\textbf{Base Classes}} & \multicolumn{2}{c|}{\textbf{Novel Classes}} & \multicolumn{2}{c|}{\textbf{All Classes}} & \multicolumn{2}{c|}{\textbf{Base Classes}} & \multicolumn{2}{c}{\textbf{Novel Classes}} \\  
 & & & 
  \textbf{AP} &
  \textbf{AP$_{50}$} &
  \textbf{AP} & 
  \textbf{AP$_{50}$} &  
  \textbf{AP} &
  \textbf{AP$_{50}$} &
  \textbf{AP} &
  \textbf{AP$_{50}$} &
  \textbf{AP} &
  \textbf{AP$_{50}$} &
  \textbf{AP} &
  \textbf{AP$_{50}$} \\
\midrule
\multirow{5}{*}{1} 
 & MRCN+ft-full~\shortcite{He-ICCV2017} & ICCV 2017 & 10.21 & 21.58 & 17.63 & 26.32 & 0.74 & 2.33 & 9.88 & 19.25 & 15.57 & 24.18 & 0.64 & 2.14 \\
 & MTFA~\shortcite{Ganea-CVPR2021} & CVPR 2021  & 24.32 & 39.64 & 31.73 & 51.49 & 2.10 & 4.07 & 22.98 & 37.48 & 29.85 & 48.64 & 2.34 & 3.99\\
 & iMTFA~\shortcite{Ganea-CVPR2021} & CVPR 2021  & 21.67 & 31.55 & 27.81 & 40.11 & 3.23 & 5.89 & 20.13 & 30.64 & 25.9 & 39.28 & 2.81 & 4.72\\
 & iFS-RCNN~\shortcite{Nguyen-CVPR2022} & CVPR 2022 & 31.19 & 52.83 & 40.08 & 71.14 & 4.54 & 10.29 & 28.45 & 46.72 & 36.35 & 63.11 & 3.95 & 7.89 \\
\rowcolor{lightgray} & \textbf{MaskDiff (Ours)} & - &  \textbf{32.59} & \textbf{54.61} & \textbf{41.23} & \textbf{72.85} & \textbf{5.26} & \textbf{11.35} & \textbf{29.42} & \textbf{48.37} & \textbf{37.59} & \textbf{64.26} & \textbf{4.85} & \textbf{8.73}\\
\midrule
\multirow{5}{*}{5} 
 & MRCN+ft-full~\shortcite{He-ICCV2017} & ICCV 2017 &  12.31 & 23.69 & 19.43 & 28.12 & 1.15 & 2.09 & 11.28 & 22.36 & 17.89 & 26.78 & 1.17 & 2.58 \\
 & MTFA~\shortcite{Ganea-CVPR2021} & CVPR 2021 &  26.39 & 41.52 & 33.11 & 51.49 & 6.22 & 11.63 & 25.07 & 39.95 & 31.29 & 49.55 & 6.38 & 11.14\\
 & iMTFA~\shortcite{Ganea-CVPR2021} & CVPR 2021 &  19.62 & 28.06 & 24.13 & 33.69 & 6.07 & 11.15 & 18.22 & 27.10 & 22.56 & 33.25 & 5.19 & 8.65\\
 & iFS-RCNN~\shortcite{Nguyen-CVPR2022} & CVPR 2022 &  32.52 & 54.30 & 40.06 & 71.19 & 9.91 & 19.24 & 29.89 & 48.22 & 36.33 & 62.81 & 8.80 & 15.73 \\
\rowcolor{lightgray} & \textbf{MaskDiff (Ours)} & - &  \textbf{33.45} & \textbf{56.19} & \textbf{41.51} & \textbf{72.28} & \textbf{10.49} & \textbf{21.07} & \textbf{30.48} & \textbf{50.35} & \textbf{38.12} & \textbf{64.36} & \textbf{9.43} & \textbf{17.12} \\
\midrule
\multirow{5}{*}{10} 
 & MRCN+ft-full~\shortcite{He-ICCV2017} & ICCV 2017 &  12.44 & 24.39 & 20.57 & 29.72 & 2.33 & 5.64 & 12.15 & 23.29 & 18.08 & 27.53 & 1.86 & 4.25 \\
 & MTFA~\shortcite{Ganea-CVPR2021} & CVPR 2021 &  27.44 & 42.84 & 33.83 & 52.04 & 8.28 & 15.25 & 25.97 & 41.28 & 31.84 & 50.17 & 8.36 & 14.58 \\
 & iMTFA~\shortcite{Ganea-CVPR2021} & CVPR 2021 &  19.26 & 27.49 & 23.36 & 32.41 & 6.97 & 12.72 & 17.87 & 26.46 & 21.87 & 32.01 & 5.88 & 9.81\\
 & iFS-RCNN~\shortcite{Nguyen-CVPR2022} & CVPR 2022 &  33.02 & 56.15 & 40.05 & 69.84 & 12.55 & 25.97 & 30.41 & 49.54 & 36.32 & 63.29 & 10.06 & 19.72\\
\rowcolor{lightgray} & \textbf{MaskDiff (Ours)} & - &  \textbf{35.21} & \textbf{59.80} & \textbf{42.17} & \textbf{72.36} & \textbf{14.04} & \textbf{28.33} & \textbf{31.89} & \textbf{52.15} & \textbf{38.55} & \textbf{66.48} & \textbf{11.84} & \textbf{21.27} \\
\bottomrule
\end{tabular}
}
\vspace{-2mm}
\caption{FSOD and FSIS performance on COCO dataset for both base and novel classes (COCO-All). MaskDiff outperforms state-of-the-art methods. The best performance is marked in boldface.}
\label{table:base_novel}
\vspace{-5mm}
\end{table*}

Classifier guidance~\cite{Dhariwal-NIPS2021} adjusts the adversarial gradient of the noisy classifier by a $\omega$ hyper-parameter term to introduce fine-grained control to either encourage or dissuade the model from accepting the conditioning information:
\begin{equation}
\resizebox{0.9\hsize}{!}{$
\nabla \log p\left(\mathbf{y}_t|\mathbf{x},\mathbf{k}, \mathbf{c}\right)
=\nabla \log p\left(\mathbf{y}_t| \mathbf{x},\mathbf{k}\right)+\omega\nabla \log p\left(\mathbf{c} | \mathbf{y}_t, \mathbf{x},\mathbf{k}\right).
$} \label{eq:cls_gamma}
\end{equation}

Inspired by classifier-free guidance~\cite{Ho-NIPSW2021}, we substitute the adversarial gradient term in Eq.~\ref{eq:cls_guide} into Eq.~\ref{eq:cls_gamma}, resulting:
\begin{equation}
\resizebox{\hsize}{!}{$
\nabla \log p\left(\mathbf{y}_t|\mathbf{x},\mathbf{k}, \mathbf{c}\right) = \underbrace{\omega\nabla \log p\left(\mathbf{y}_t| \mathbf{x},\mathbf{k}, \mathbf{c}\right)}_{\text {guidance-specific score }}+\underbrace{(1-\omega)\nabla \log p\left(\mathbf{y}_t|\mathbf{x},\mathbf{k}\right)}_{\text {guidance-agnostic score}}.$}
\end{equation}

As $\nabla \log p\left(\mathbf{y}_t|\mathbf{x},\mathbf{k}, \mathbf{c}\right) = -\frac{1}{\sqrt{1-\bar{\alpha}_t}}\boldsymbol{\epsilon}_\theta(\mathbf{y}_t, \mathbf{x}, \mathbf{k}, \mathbf{c}, t)$ (see Supplementary Materials for more details), we have the equivalent form:
\begin{equation}
\resizebox{\hsize}{!}{$
\boldsymbol{\hat{\epsilon}}_\theta(\mathbf{y}_t, \mathbf{x}, \mathbf{k}, \mathbf{c}, t) = \underbrace{\omega\boldsymbol{\epsilon}_\theta(\mathbf{y}_t, \mathbf{x}, \mathbf{k}, \mathbf{c}, t)}_{\text {guidance-specific score }}+\underbrace{(1-\omega)\boldsymbol{\epsilon}_\theta(\mathbf{y}_t, \mathbf{x}, \mathbf{k}, t)}_{\text {guidance-agnostic score}}.$}
\end{equation}

The inference procedure with guided sampling is shown in Alg.~\ref{alg:inference}. When $\omega > 1$, the diffusion model not only favors the guidance-specific score function over the guidance-agnostic one but also moves away from it. Along with producing samples that accurately match the conditioning information, this also reduces sample variety and makes generated masks more stable. Following Ho~\etal~\shortcite{Ho-NIPSW2021}, we set $\omega = 5$ to obtain the most individual sample fidelity.

\section{Experiments}
\begin{table}[!t]
\centering
\footnotesize
\resizebox{1\linewidth}{!}{
\begin{tabular}{c|l|cc|cc}
\toprule
\multirow{2}{*}{\textbf{Shots}} & \multirow{2}{*}{\textbf{Method}} & \multicolumn{2}{c|}{\textbf{Detection}} & \multicolumn{2}{c}{\textbf{Segmentation}} \\ 
 &
 &
  \textbf{AP} &
  \textbf{AP$_{50}$} &
  \textbf{AP} & 
  \textbf{AP$_{50}$}  \\
\midrule
\multirow{6}{*}{10} & MRCN+ft-full~\shortcite{He-ICCV2017} & 2.52 & 5.78  & 1.93 & 4.68  \\
& Meta-RCNN~\shortcite{Yan-ICCV2019} & 5.60 & 14.20 & 4.40 & 10.60 \\
 & MTFA~\shortcite{Ganea-CVPR2021} & 8.52 & 15.53 & 8.40 & 14.62  \\
 & iMTFA~\shortcite{Ganea-CVPR2021} & 7.14 & 12.91 &  5.94 & 9.96  \\
 & iFS-RCNN~\shortcite{Nguyen-CVPR2022} & 11.27 & 22.15  & 10.22 & 20.61  \\
\rowcolor{lightgray} & \textbf{MaskDiff (Ours)} & \textbf{12.18} & \textbf{23.61} & \textbf{11.01} & \textbf{21.58} \\
\bottomrule
\end{tabular}
}
\caption{FSOD and FSIS performance on only COCO novel classes (COCO-Novel). MaskDiff outperforms state-of-the-art methods. Results on other $K-$shot are shown in Supp.
}
\label{table:novel}
\end{table}

\subsection{Implementation Details}
Our MaskDiff was implemented using the Detectron2 library
. The used backbone architecture was a ResNet50~\cite{He-CVPR2016}, and we utilized FPN~\cite{Lin-CVPR2017} for the path aggregation block. The mask distribution modeling head was implemented using Guided Diffusion~\cite{Dhariwal-NIPS2021,Ho-NIPSW2021} with $1,000$ reverse steps. In the base training stage, similar to TFA~\cite{Wang-ICML2020}, object detector heads were trained using SGD with a learning rate of $0.02$, momentum of $0.9$, and weight decay of $10^{-4}$. The mask distribution modeling head was trained using AdamW with a learning rate of $10^{-4}$, $\beta_1=0.9$, and $\beta_2=0.999$. We also used an exponential moving average (EMA) rate of $0.9999$. Object regions, binary masks, and $K-$shot guidance were all resized to $128\times128$ for training the mask distribution modeling head. In the few-shot fine-tuning stage, the entire network, including object detector heads and the mask distribution modeling head, was jointly trained with the same configuration as training the mask distribution modeling head in the first stage. In the final step of the denoising process, we set the threshold of noisy binary masks to $0.5$ to separate $0$ and $1$ values. All variants of MaskDiff were trained with a batch size of $8$ on a single NVIDIA RTX 3090 GPU.
\subsection{Experimental Settings}
Following the standard FSIS evaluation~\cite{Wang-ICML2020,Ganea-CVPR2021}, we evaluated MaskDiff on COCO~\cite{Lin-ECCV2014}, VOC2007 and VOC2012~\cite{Everingham-IJCV2010} datasets. $80$ classes of COCO were split as suggested by Kang~\etal~\shortcite{Kang-ICCV2019}. $20$ classes that intersect with VOC were used as novel classes, whereas the remaining $60$ classes were used as base classes. The union of COCO training set ($80k$ images) and COCO validation set ($35k$ images) was utilized for training, while the remaining $5k$ images served as the test set. We also combined validation sets of VOC2007 and VOC2012 only for testing.

\begin{table}[!t]
\centering
\footnotesize
\resizebox{1\linewidth}{!}{
\begin{tabular}{c|l|cc|cc}
\toprule
\multirow{2}{*}{\textbf{Shots}} & \multirow{2}{*}{\textbf{Method}} & \multicolumn{2}{c|}{\textbf{Detection}} & \multicolumn{2}{c}{\textbf{Segmentation}} \\ 
 &
 &
  \textbf{AP} &
  \textbf{AP$_{50}$} &
  \textbf{AP} & 
  \textbf{AP$_{50}$}  \\
\midrule
\multirow{4}{*}{1}
 & FGN~\shortcite{Fan-CVPR2020} & - & 30.80 & - & 16.20 \\
 & MTFA~\shortcite{Ganea-CVPR2021} & 9.99 & 21.68 & 9.51 & 19.28 \\
 & iMTFA~\shortcite{Ganea-CVPR2021} & 11.47 & 22.41 & 8.57 & 16.32 \\
\rowcolor{lightgray} & \textbf{MaskDiff (Ours)} & \textbf{24.15} & \textbf{38.57} & \textbf{22.73} & \textbf{37.59} \\
\bottomrule
\end{tabular}
}
\vspace{-2mm}
\caption{FSIS performance on COCO2VOC. MaskDiff outperforms state-of-the-art methods. The authors of FGN~\cite{Fan-CVPR2020} reported only AP50 results of one-shot.}
\label{table:cross_dataset}
\vspace{-6mm}
\end{table}

\begin{figure*}[t!]
    \centering
    \includegraphics[width=\textwidth]{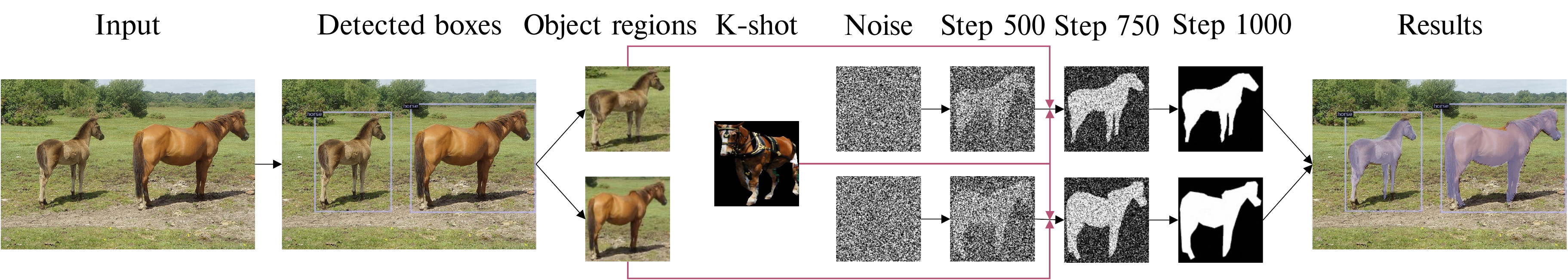}
    \vspace{-4mm}
    \caption{ 
    Inference example of our MaskDiff when training and inference on one-shot setting for the COCO novel classes. }
    \label{fig:qualitative}
    \vspace{-5mm}
\end{figure*}
We compared the performance of $K = 1, 5, 10$ shots per novel class. We repeated all tests $10$ times with $K$ random samples per class to limit the influence of outliers caused by the random selection of $K$ shots and report the mean result. Our FSIS assessment approach is identical to that of few-shot object detection (FSOD)~\cite{Wang-ICML2020}. Crucially, segmentation results of different runs are not unique since the masks sampled in the inference depend on random factors ($\mathbf{y}_T \sim \mathcal{N}(\mathbf{0}, \mathbf{I})$, $\boldsymbol{\epsilon} \sim \mathcal{N}(\mathbf{0}, \mathbf{I})$). To provide a stable and reliable evaluation procedure of generative-based models, we propose a simple yet effective evaluation inspired by TFA~\cite{Wang-ICML2020}. With a model trained on specific $K$ shots, we performed inference for multiple runs on different random seeds to obtain averages and confidence intervals. We observed that $10$ runs are comparatively reliable and stable for evaluation reports. To sum up, we trained models for $10$ runs on different samples of training shots, and with each trained model, we tested on $10$ different random seeds. 

\subsection{Comparison with State-of-the-art Methods}
\textbf{Results on both COCO base and novel classes (COCO-All).} In this experiment, we aim to predict all $80$ COCO classes and report the standard evaluation metrics, including mAP and AP50. Following the newly introduced evaluation process of Ganea~\etal~\shortcite{Ganea-CVPR2021}, we report the performance of the base and novel classes independently as well as all classes. We compared our MaskDiff against state-of-the-art methods, such as MTFA~\cite{Ganea-CVPR2021}, iMTFA~\cite{Ganea-CVPR2021}, iFS-RCNN~\cite{Nguyen-CVPR2022}, and a fully-converged Mask R-CNN model fine-tuned on the novel classes (MRCN+ft-full)~\cite{He-ICCV2017}). Results in Tab.~\ref{table:base_novel} show that MaskDiff consistently outperforms state-of-the-art methods on both FSOD and FSIS tasks for all numbers of provided shots on both base and novel classes. This implies MaskDiff can adapt to novel classes while embracing performance in base classes. Especially, MaskDiff outperforms MTFA and iMTFA, the two SOTA methods utilizing TFA~\cite{Wang-ICML2020} as the core of developed networks, with significant margins on the COCO dataset. Particularly, compared with MTFA, for $10$ shots, our performance of FSIS gains is about $+3.5$ on new classes and $+7$ on base classes. Likewise, our MaskDiff also dramatically outperforms iMTFA with gains of $+6$ on novel classes and $+17$ on base classes. Modeling mask distribution via DPM is more effective than conventional methods and gains large performance in learning existing and new categories on limited data. Regarding FSOD, our MaskDiff is built upon TFA, a multi-task learning design, i.e., detection and segmentation. The improvement in one task can benefit another one. As our segmentation head generates a better binary mask for each region. This information can be used to improve the accuracy of object detection by providing a more precise location and shape of the object. 

\begin{table}[t!]
\centering
\resizebox{\linewidth}{!}{
\begin{tabular}{l|cc|c}
\midrule
 \multirow{2}{*}{\textbf{Method}} & \multicolumn{2}{c|}{\textbf{Training Time (Hours)}} & \multirow{2}{*}{\textbf{Inference Time (FPS)}} \\ 
 &
  \textbf{Base} &
  \textbf{Fine-tuning} & 
 \\
\midrule
 MRCN+ft-full~\shortcite{He-ICCV2017} & 24.6 & 12.8 & 10.38 \\
 MTFA~\shortcite{Ganea-CVPR2021} & 24.8 & 12.7  & 10.32 \\
 iMTFA~\shortcite{Ganea-CVPR2021} & \textbf{24.2} & \textbf{12.4} & \textbf{11.75} \\
 iFS-RCNN~\shortcite{Nguyen-CVPR2022} & 28.5 & 15.6 & 10.28 \\
 \textbf{MaskDiff (Ours)} & 32.3 & 16.1 & 8.27\\
\midrule
\end{tabular}}
\vspace{-2mm}
\caption{Runtime of FSIS ($1-$shot) on COCO dataset.}
\label{table:running_time}
\vspace{-6mm}
\end{table}

\textbf{Results on only COCO novel classes (COCO-Novel).} Table~\ref{table:novel} reports our results on only COCO novel classes, in which we compared MaskDiff against SOTA FSIS methods (\eg, Meta-RCNN~\cite{Yan-ICCV2019}, MTFA~\cite{Ganea-CVPR2021}, iMTFA~\cite{Ganea-CVPR2021}, iFS-RCNN~\cite{Nguyen-CVPR2022}, and MRCN+ft-full~\cite{He-ICCV2017}). Experimental results demonstrated the superior performance of MaskDiff on both detection and segmentation tasks consistently on all $K-$shots. MaskDiff surpasses the recent iFS-RCNN~\cite{Nguyen-CVPR2022} and drastically outperforms other methods by a large margin in terms of AP. This indicates that our proposal is better than previous strategies including episodic-training (Meta-RCNN~\shortcite{Yan-ICCV2019}), class-specific mask predictor (MTFA~\shortcite{Ganea-CVPR2021}), and class-agnostic one (iMTFA~\shortcite{Ganea-CVPR2021}, iFS-RCNN~\shortcite{Nguyen-CVPR2022}).

\begin{table}[!t]
\centering
\resizebox{1\columnwidth}{!}{
\begin{tabular}{c|cc|cc|cc}
\toprule
\multirow{2}{*}{\textbf{Shots}} & \multicolumn{2}{c|}{\textbf{MaskDiff}} & \multicolumn{2}{c|}{\textbf{Detection}} & \multicolumn{2}{c}{\textbf{Segmentation}} \\ 
 &
 Guided & Two-Stage Training
 &
  \textbf{AP} &
  \textbf{AP$_{50}$} &
  \textbf{AP} & 
  \textbf{AP$_{50}$} \\
\midrule
\multirow{4}{*}{10}
 & \ding{55} & \ding{55} & 10.59 & 20.96 & 9.68 & 18.83 \\
 & \ding{51} & \ding{55} & 10.86 & 21.55 &  10.02 & 19.48  \\
 & \ding{55} & \ding{51} & 12.05 & 23.32  & 10.76 & 20.63  \\
\rowcolor{lightgray}
 & \ding{51} & \ding{51} & 12.18 & 23.61 & 11.01 & 21.58  \\
\bottomrule
\end{tabular}
}
\vspace{-2mm}
\caption{Our ablation study on COCO-Novel dataset. MaskDiff with classifier-free guided mask sampling and diffusion-based two-stage training strategy performs best. Results on other $K-$shot are shown in Supp.}
\label{table:ablation}
\vspace{-5mm}
\end{table}

\textbf{Cross-dataset COCO2VOC evaluation.} To investigate the ability to learn new categories, we compared MaskDiff with some SOTA FSIS methods (\eg, FGN~\cite{Fan-CVPR2020}, MTFA~\cite{Ganea-CVPR2021}, and iMTFA~\cite{Ganea-CVPR2021}) on cross-dataset COCO2VOC setting. Methods were trained on COCO dataset and evaluated on VOC dataset. We reported public results provided by authors, which are only in one-shot setting ($K=1$). Table~\ref{table:cross_dataset} shows that our MaskDiff transcends cutting-edge methods on cross-dataset evaluation. More importantly, our mask distribution modeling approach outperforms point estimation one~\cite{Fan-CVPR2020}.


\textbf{Qualitative evaluation.} Figure~\ref{fig:qualitative} demonstrates an inference example of our MaskDiff when training and inference on the one-shot setting for the COCO novel classes. More successful and failed cases can be found in Supp.

\textbf{Runtime evaluation.} Regarding the training and inference time, Table~\ref{table:running_time} shows that other methods can be trained and can infer faster than ours. However, we remark that for few-shot learning tasks, effectiveness (AP) should be prioritized rather than efficiency (processing time).

\subsection{Ablation Study}

\begin{table}[!t]
\centering
\resizebox{0.9\columnwidth}{!}{
\begin{tabular}{c|c|cc}
\toprule
\multirow{2}{*}{\textbf{Shots}} & \multirow{2}{*}{\textbf{MaskDiff}} &
\multicolumn{2}{c}{\textbf{Segmentation}} \\ 
 &
 &
  \textbf{AP} & 
  \textbf{AP$_{50}$} \\
\multirow{2}{*}{1} 
 &  w/o guided sampling  & 6.04 $\pm$ 0.09 & 12.29 $\pm$ 0.12 \\
 &  \textbf{w/ guided sampling}  & 6.23 $\pm$ \textbf{0.07} & 12.47 $\pm$ \textbf{0.08} \\
\bottomrule
\end{tabular}
}
\vspace{-2mm}
\caption{We train our MaskDiff on a specific set of training samples and perform inference on 10 different random seeds. Then, we compute the mean and standard deviation (std) AP of those runs. MaskDiff with guided sampling not only outperforms the one without guidance but also is more stable (less std). Qualitative results of classifier-free guided mask sampling and other $K-$shot can be seen in Supp.}
\label{table:significance_cls}
\vspace{-2mm}
\end{table}

\begin{table}[t!]
\centering
\resizebox{\columnwidth}{!}{
\begin{tabular}{l|c|c|c}
\toprule
\textbf{Method} & \textbf{All Classes} & \textbf{Base Classes} & \textbf{Novel Classes} \\  
 MaskDiff w/ basic UNet  & 24.16 $\pm$ 0.22 & 31.49 $\pm$ 0.33 & 3.32 $\pm$ 0.35\\
  \textbf{MaskDiff w/ diffusion} & \textbf{29.42} \textbf{{$\pm$ 0.09}} & \textbf{37.59} \textbf{{$\pm$ 0.0093}} & \textbf{4.85} \textbf{{$\pm$ 0.24}} \\
\bottomrule
\end{tabular}
}
\vspace{-2mm}
\caption{\small{Stability and effectiveness of our MaskDiff with and without diffusion on COCO-All dataset ($1-$shot).}}
\label{table:maskdiff_unet}
\vspace{-5mm}
\end{table}

\textbf{Effectiveness of two-stage training strategy.} We aim to evaluate the effectiveness of the proposed two-stage training strategy consisting of base training and few-shot fine-tuning. In base training, we first train the object detector and then $K-$shot mask modeling on base images. In the fine-tuning stage, we jointly train both object detector heads and our mask distribution modeling head on balanced base and novel classes. We compared our approach with the common approach using only a few-shot fine-tuning stage. Particularly, we exclude the first stage training on base images (\ie, base training) and utilize only the second stage. In early steps, the object detector cannot precisely localize and classify the objects. Given unsatisfactory localized object regions, the mask distribution modeling head is much more difficult to train, leading to generating not good segmentation masks. Table~\ref{table:ablation} shows a considerable collapse in results. 

\textbf{Significance of classifier-free guided mask sampling.} We studied the effectiveness as well as stability of classifier-free guided mask sampling by comparing the performance MaskDiff with and without classifier-free guided mask sampling. From Table~\ref{table:ablation}, we conclude that classifier-free guided mask sampling boosts the performance of FSIS. Furthermore, we evaluated the standard deviation of testing results on 10 different random seeds when running on a specific sample of training shots. In Table~\ref{table:significance_cls}, we can see that the standard deviation of MaskDiff with guidance over 10 different random seeds is less than that of without guidance. In other words, the classifier-free guided mask sampling makes the mask distribution modeling head more stable. 
\begin{figure}[!t]
    \centering
    \includegraphics[width=\linewidth]{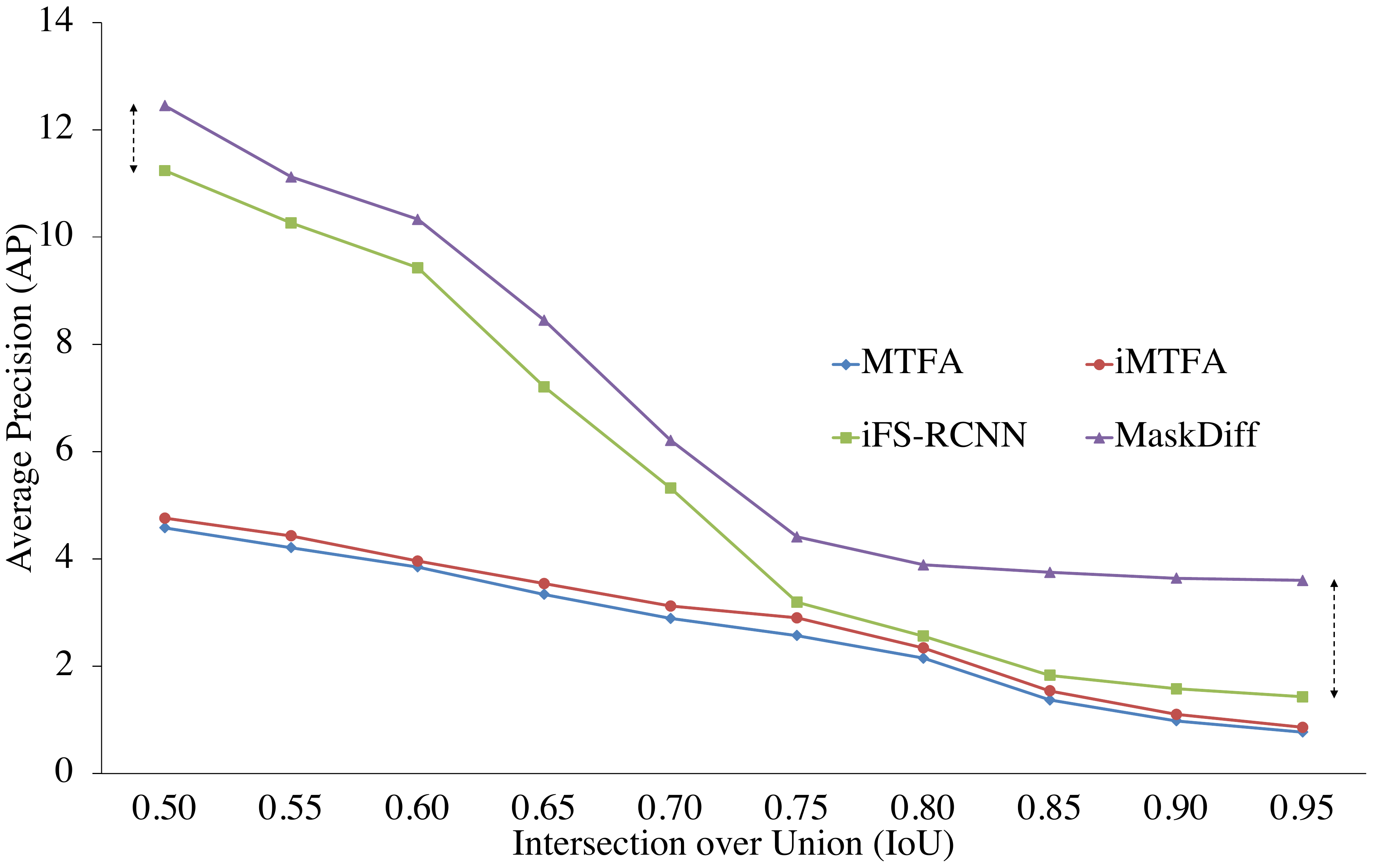}
    \caption{MaskDiff preserves spatial information, especially at detailed levels. We compare the performance of MaskDiff with state-of-the-art methods in one-shot instance segmentation at different IoU thresholds. MaskDiff outperforms other methods with large margins at high IoU thresholds, which indicates its ability to segment objects more precisely.}    
    \label{fig:iou-1shot}
    \vspace{-2mm}
\end{figure}

\textbf{Effectiveness of diffusion probabilistic model.} We replace our diffusion with the same UNet (without noise) to verify the effectiveness of our diffusion head. Table~\ref{table:maskdiff_unet} shows that replacing our diffusion with the basic UNet downgrades the effectiveness as well as stability of our method significantly, consolidating the contribution of our mask distribution modeling approach via DPM.

\textbf{Stability analysis.} The performance of various methods has been generally evaluated based on the evaluation protocol proposed by Wang~\etal~\shortcite{Wang-ICML2020}. In particular, a model is trained on a different set of training samples and is evaluated on the same test set to report the mean result. We further computed the standard deviation of different runs and compared the stability with state-of-the-art methods. Table~\ref{table:stability} shows that MaskDiff achieves the highest performance and the lowest standard deviation in terms of AP, indicating MaskDiff is the most stable and robust to different $K-$shot. 

\begin{table}[t!]
\centering
\resizebox{1\columnwidth}{!}{
\begin{tabular}{c|l|c|c|c}
\toprule
\textbf{Shots} & \textbf{Method} & \textbf{All Classes} & \textbf{Base Classes} & \textbf{Novel Classes} \\  
\midrule
\multirow{5}{*}{1} & MRCN+ft-full~\shortcite{He-ICCV2017} & 9.88 $\pm$ 0.31  & 15.57 $\pm$ 0.30 & 0.64 $\pm$ 0.26 \\
 & MTFA~\shortcite{Ganea-CVPR2021} & 22.98 $\pm$ 0.24 & 29.85 $\pm$ 0.35 & 2.34 $\pm$ 0.31 \\
 & iMTFA~\shortcite{Ganea-CVPR2021} & 20.13 $\pm$ 0.28 & 25.90 $\pm$ 0.32 & 2.81 $\pm$ 0.37 \\
 & iFS-RCNN~\shortcite{Nguyen-CVPR2022}  & 28.45 $\pm$ 0.12 & 36.35 $\pm$ 0.01 &  3.95 $\pm$ 0.48\\
\rowcolor{lightgray} & \textbf{MaskDiff (Ours)} & 29.42 \textbf{{$\pm$ 0.09}} & 37.59  \textbf{{$\pm$ 0.0093}} & 4.85 \textbf{{$\pm$ 0.24}}\\
\bottomrule
\end{tabular}
}
\vspace{-1mm}
\caption{Stability of FSIS on COCO-All dataset. MaskDiff is the most stable method with the smallest standard deviation of AP. The best performance is marked in boldface. Results on other $K-$shot are described in Supp.}
\label{table:stability}
\vspace{-5mm}
\end{table}

\textbf{Spatial information preservation.} Most conventional FSIS methods leverage pooled object features and feed them into FCN~\cite{Long-CVPR2015} to generate binary masks. However, passing through several pooling layers leads to a severe collapse in spatial information, especially the detailed one. By directly utilizing object regions from input images to generate binary masks, our MaskDiff can embrace comprehensive spatial details and induce precise delineation. In Fig.~\ref{fig:iou-1shot}, we visualize the line graph representing AP from $0.50$ to $0.95$ with step size $0.05$ to illustrate the changes in the performance of FSIS methods when the IoU requirement rises. Our MaskDiff surpasses cutting-edge methods, especially at high IoU thresholds.

\section{Conclusion}
We presented the first mask distribution modeling approach for FSIS and designed a novel method dubbed MaskDiff. We adapted conditional DPM to model binary mask distribution conditioned on RGB object regions and $K-$shot samples. Furthermore, we leveraged the off-the-shelf classifier branch to guide the mask sampling procedure with the category information. Experimental results demonstrated that our proposed method achieves state-of-the-art results while being more stable than existing methods. 

\section{Acknowledgments} 
This research was funded by Vingroup and supported by Vingroup Innovation Foundation (VINIF) under project code VINIF.2019.DA19. The project was also supported by National Science Foundation Grant (NSF\#2025234).

\bibliography{aaai24}

\end{document}


\maketitle

In this supplementary material, we provide a detailed and 
comprehensive description of our proposed method and experimental results. In section~\ref{sec:architecture}, we present the denoising architecture of the mask distribution modeling branch. Section~\ref{sec:derivation} provides the detailed derivation of diffusion probabilistic model to estimate the conditional distribution of a binary mask conditioned on an object region and $K-$shot information. In section~\ref{sec:experiments}, we demonstrate comprehensive experiments in detail including comparison with other methods, results on $K=1,5,10$ shots, extended ablation study and qualitative evaluation.

\section{Denoising Architecture}
\begin{figure}[h]
    \centering
    \includegraphics[width=\linewidth]{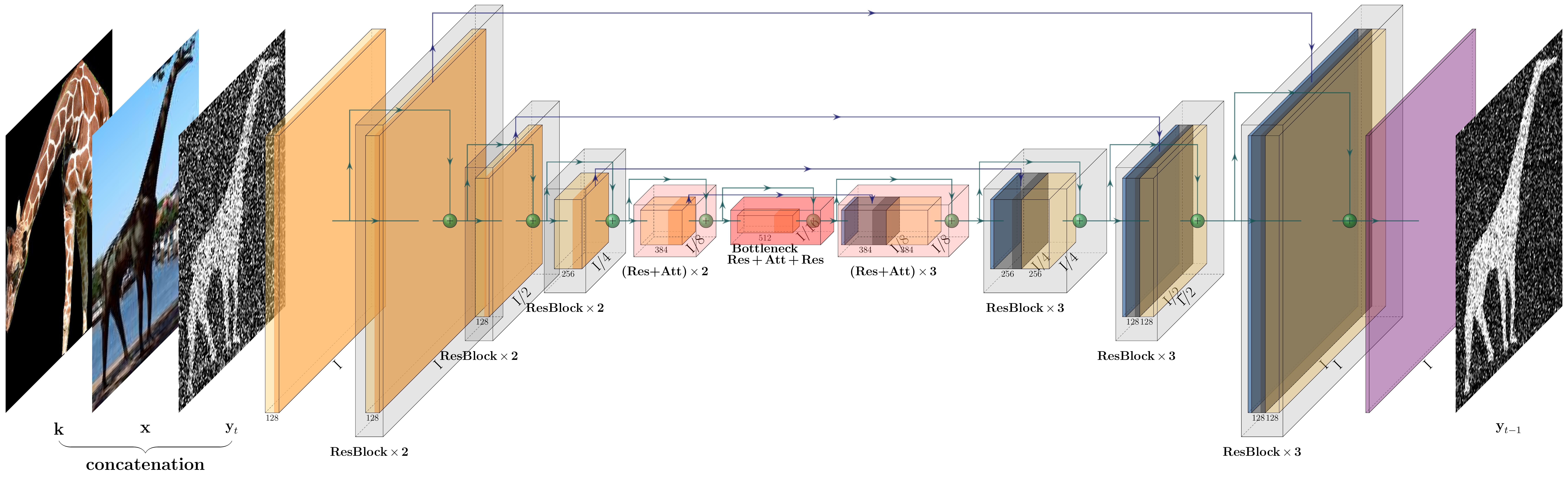}
    \caption{Denoising architecture of diffusion probabilistic model for mask distribution modeling. This architecture is based on the UNet architecture with additional modifications including addding Residual Block (ResBlock) and Attention Block (Att). With respect to the conditioning module, we simply concatenate three components namely $\mathbf{y}_t,\mathbf{x}, \mathbf{k}$ and feed it into the network for generating the less noisy version $\mathbf{y}_{t-1}$.}    
    \label{fig:denoising_arch}
\end{figure}

\setlabel{Denoising Architecture}{sec:architecture}
In this section, we visualize the architecture of denoising head of the conditional diffusion probabilistic model (see Fig.~\ref{fig:denoising_arch}). In particular, the denoising head follows UNet architecture with some adjustments. Residual blocks are utilized at every feature resolution. In the downsample branch, two residual blocks are applied at each resolution while three residual ones are used in the upsample branch. Furthermore, attention blocks are leveraged at three parts (the bottleneck features and the features with 384 channels in both downsample and upsample branches, attention blocks (Att) are in red color). Regarding the conditioning module, we simply concatenate three components including $\mathbf{y}_t,\mathbf{x},$ and $\mathbf{k}$. Then we feed the concatenated input into the denoising network to generate the less noisy version of the binary mask $\mathbf{y}_{t-1}$. Likewise, in $K > 1$ cases, we also simply concatenate all of these provided information. 

\textbf{Guidance-specific score.} We used the same architecture and added an additional one-hot class vector. At each Residual Block, the feature has the shape of $(B, C, H, W)$. We embed the one-hot vector of size $N$ (number of classes) to the same dimension $C$, resizing it to the shape of $(B, C, 1, 1)$ and then plus them with the corresponding feature at each Residual Block.

\section{Detailed Derivations}
\setlabel{Detailed Derivations}{sec:derivation}
\subsection{Forward Diffusion Process}
The forward diffusion process is defined gradually adding a small amount of Gaussian noise to the sample in $T$ steps, producing a sequence of noisy samples $\mathbf{y}_1, \dots,$ and $\mathbf{y}_T$, formulated as follows:
\begin{equation}
    q\left(\mathbf{y}_{1: T} \vert \mathbf{y}_{0}\right)=\prod_{t=1}^{T} q\left(\mathbf{y}_{t} \vert \mathbf{y}_{t-1}\right),
\end{equation}
where Gaussian noise is added at each iteration of the forward diffusion process as follows:
\begin{equation}
    q\left(\mathbf{y}_{t} \vert \mathbf{y}_{t-1}\right)=\mathcal{N}\left(\mathbf{y}_{t} ; \sqrt{1-\beta_{t}} \mathbf{y}_{t-1}, \beta_{t} \mathbf{I}\right).
\end{equation}

The data sample $\mathbf{y}_0$ gradually loses its distinguishable features as step $t$ increases. Eventually, when $T \rightarrow \infty, \mathbf{y}_T$ is equivalent to an isotropic Gaussian distribution. The step sizes are controlled by a variance schedule $\left\{\beta_{t} \in(0,1)\right\}_{t=1}^{T}$.
An important property of the forward process is that we can sample $\mathbf{y}_t$ at any arbitrary time step $t$ in a closed form using the reparameterization trick. Let $\alpha_{t}=1-\beta_{t}$ and $\bar{\alpha}_{t}=\prod_{i=1}^{T} \alpha_i$, we have:
\begin{equation}
\begin{split}
\mathbf{y}_{t} & =\sqrt{\alpha_{t}} \mathbf{y}_{t-1}+\sqrt{1-\alpha_{t}} \boldsymbol{\epsilon}_{t-1} \quad ; \text { where } \boldsymbol{\epsilon}_{t-1}, \boldsymbol{\epsilon}_{t-2}, \cdots \sim \mathcal{N}(\mathbf{0}, \mathbf{I}) \\
& =\sqrt{\alpha_{t} \alpha_{t-1}} \mathbf{y}_{t-2}+\sqrt{1-\alpha_{t} \alpha_{t-1}} \overline{\boldsymbol{\epsilon}}_{t-2} \quad ; \text { where } \overline{\boldsymbol{\epsilon}}_{t-2} \text { merges two Gaussians } \\
& =\cdots \\
& =\sqrt{\bar{\alpha}_{t}} \mathbf{y}_{0}+\sqrt{1-\bar{\alpha}_{t}} \boldsymbol{\epsilon}. 
\end{split}
\label{eq:yt_y0_repara}
\end{equation}

Based on the above analysis, we obtain:
\begin{equation} 
\label{eq:yt_y0}
q\left(\mathbf{y}_{t} \vert \mathbf{y}_{0}\right) =\mathcal{N}\left(\mathbf{y}_{t} ; \sqrt{\bar{\alpha}_{t}} \mathbf{y}_{0},\left(1-\bar{\alpha}_{t}\right) \mathbf{I}\right).
\end{equation}

\subsection{Reverse Diffusion Process}
The reverse diffusion process $p_{\theta}\left(\mathbf{y}_{0: T} \vert \mathbf{x}, \mathbf{k}\right)$ is defined as a Markov chain with learned Gaussian transitions beginning with $p(\mathbf{y}_T) \sim \mathcal{N}(\mathbf{0}, \mathbf{I})$, which is formulated as follows,
\begin{equation}
p_{\theta}\left(\mathbf{y}_{0: T} \vert \mathbf{x}, \mathbf{k}\right)=p\left(\mathbf{y}_{T}\right) \prod_{t=1}^{T} p_{\theta}\left(\mathbf{y}_{t-1} \vert \mathbf{y}_{t}, \mathbf{x}, \mathbf{k}\right),
\end{equation}
where reverse process steps are performed by taking small Gaussian steps described by:
\begin{equation}
p_{\theta}\left(\mathbf{y}_{t-1} \vert \mathbf{y}_{t}, \mathbf{x},\mathbf{k}\right)=\mathcal{N}\left(\mathbf{y}_{t-1} ; \boldsymbol{\mu}_{\theta}\left(\mathbf{y}_{t}, \mathbf{x}, \mathbf{k}, t\right), \mathbf{\Sigma}_{\theta}\left(\mathbf{y}_{t}, \mathbf{x}, \mathbf{k}, t\right)\right).    
\end{equation}

It is essential to notice that the reverse conditional probability is tractable when conditioned on $\mathbf{y}_0$:
\begin{equation}
    q(\mathbf{y}_{t-1} \vert \mathbf{y}_t, \mathbf{y}_0)= \mathcal{N}(\mathbf{y}_{t-1};\tilde{\boldsymbol{\mu}}(\mathbf{y}_t, \mathbf{y}_0), \tilde{\beta}_t \mathbf{I})
\end{equation}

Using Bayes' rule, we obtain:
\begin{equation}
\begin{split}
q(\mathbf{y}_{t-1} \vert \mathbf{y}_t, \mathbf{y}_0) &= q(\mathbf{y}_t \vert \mathbf{y}_{t-1}, \mathbf{y}_0) \frac{ q(\mathbf{y}_{t-1} \vert \mathbf{y}_0) }{ q(\mathbf{y}_t \vert \mathbf{y}_0) } \\
&\propto \exp \Big(-\frac{1}{2} \big(\frac{(\mathbf{y}_t - \sqrt{\alpha_t} \mathbf{y}_{t-1})^2}{\beta_t} + \frac{(\mathbf{y}_{t-1} - \sqrt{\bar{\alpha}_{t-1}} \mathbf{y}_0)^2}{1-\bar{\alpha}_{t-1}} - \frac{(\mathbf{y}_t - \sqrt{\bar{\alpha}_t} \mathbf{y}_0)^2}{1-\bar{\alpha}_t} \big) \Big) \\
&= \exp \Big(-\frac{1}{2} \big(\frac{\mathbf{y}_t^2 - 2\sqrt{\alpha_t} \mathbf{y}_t \mathbf{y}_{t-1} + \alpha_t \mathbf{y}_{t-1}^2}{\beta_t} + \frac{ \mathbf{y}_{t-1}^2 - 2 \sqrt{\bar{\alpha}_{t-1}} \mathbf{y}_0 \mathbf{y}_{t-1} + \bar{\alpha}_{t-1} \mathbf{y}_0^2  }{1-\bar{\alpha}_{t-1}} - \frac{(\mathbf{y}_t - \sqrt{\bar{\alpha}_t} \mathbf{y}_0)^2}{1-\bar{\alpha}_t} \big) \Big) \\
&= \exp\Big( -\frac{1}{2} \big((\frac{\alpha_t}{\beta_t} + \frac{1}{1 - \bar{\alpha}_{t-1}}) \mathbf{y}_{t-1}^2 - (\frac{2\sqrt{\alpha_t}}{\beta_t} \mathbf{y}_t + \frac{2\sqrt{\bar{\alpha}_{t-1}}}{1 - \bar{\alpha}_{t-1}} \mathbf{y}_0) \mathbf{y}_{t-1} + C(\mathbf{y}_t, \mathbf{y}_0) \big) \Big),
\end{split}
\end{equation}
where $C(\mathbf{y}_t, \mathbf{y}_0)$ is some function not involving $\mathbf{y}_{t-1}$ and details are omitted. Following the standard Gaussian density function, the mean and variance of $q(\mathbf{y}_{t-1} \vert \mathbf{y}_t, \mathbf{y}_0)$ can be parameterized as follows (recall that $\alpha_t = 1 - \beta_t$ and $\bar{\alpha}_t = \prod_{i=1}^T \alpha_i$):
\begin{equation}
 \tilde{\beta}_t = 1/(\frac{\alpha_t}{\beta_t} + \frac{1}{1 - \bar{\alpha}_{t-1}}) 
= 1/(\frac{\alpha_t - \bar{\alpha}_t + \beta_t}{\beta_t(1 - \bar{\alpha}_{t-1})})
= \frac{1 - \bar{\alpha}_{t-1}}{1 - \bar{\alpha}_t} \cdot \beta_t.
\end{equation}
\begin{equation}
\begin{split}
\tilde{\boldsymbol{\mu}}_t (\mathbf{y}_t, \mathbf{y}_0)
&= (\frac{\sqrt{\alpha_t}}{\beta_t} \mathbf{y}_t + \frac{\sqrt{\bar{\alpha}_{t-1} }}{1 - \bar{\alpha}_{t-1}} \mathbf{y}_0)/(\frac{\alpha_t}{\beta_t} + \frac{1}{1 - \bar{\alpha}_{t-1}}) \\
&= (\frac{\sqrt{\alpha_t}}{\beta_t} \mathbf{y}_t + \frac{\sqrt{\bar{\alpha}_{t-1} }}{1 - \bar{\alpha}_{t-1}} \mathbf{y}_0) \frac{1 - \bar{\alpha}_{t-1}}{1 - \bar{\alpha}_t} \cdot \beta_t \\
&= \frac{\sqrt{\alpha_t}(1 - \bar{\alpha}_{t-1})}{1 - \bar{\alpha}_t} \mathbf{y}_t + \frac{\sqrt{\bar{\alpha}_{t-1}}\beta_t}{1 - \bar{\alpha}_t} \mathbf{y}_0.\\
\end{split}
\label{eq:mu_}
\end{equation}

From Eq.~\ref{eq:yt_y0_repara}, we can represent $\mathbf{y}_0 = \frac{1}{\sqrt{\bar{\alpha}_t}}(\mathbf{y}_t - \sqrt{1 - \bar{\alpha}_t}\boldsymbol{\epsilon}_t)$ and plug it into the Eq.~\ref{eq:mu_} and obtain:
\begin{equation}
\begin{split}
\tilde{\boldsymbol{\mu}}_t
&= \frac{\sqrt{\alpha_t}(1 - \bar{\alpha}_{t-1})}{1 - \bar{\alpha}_t} \mathbf{y}_t + \frac{\sqrt{\bar{\alpha}_{t-1}}\beta_t}{1 - \bar{\alpha}_t} \frac{1}{\sqrt{\bar{\alpha}_t}}(\mathbf{y}_t - \sqrt{1 - \bar{\alpha}_t}\boldsymbol{\epsilon}_t) \\
&= \frac{1}{\sqrt{\alpha_t}} \Big( \mathbf{y}_t - \frac{\beta_t}{\sqrt{1 - \bar{\alpha}_t}} \boldsymbol{\epsilon}_t \Big).
\end{split}
\end{equation}
\subsection{Diffusion Loss}
The conditional diffusion probabilistic model is trained to minimize the cross entropy as the learning objective, which is equivalent to minimize variational upper bound (VUB): 
\begin{equation}
\begin{split}
L_\text{CE} &= - \mathbb{E}_{q(\mathbf{y}_0\vert \mathbf{x}, \mathbf{k})} \log p_\theta(\mathbf{y}_0\vert \mathbf{x}, \mathbf{k}) \\
&= - \mathbb{E}_{q(\mathbf{y}_0\vert \mathbf{x}, \mathbf{k})} \log \Big( \int p_\theta(\mathbf{y}_{0:T}\vert \mathbf{x}, \mathbf{k}) d\mathbf{y}_{1:T} \Big) \\
&= - \mathbb{E}_{q(\mathbf{y}_0\vert \mathbf{x}, \mathbf{k})} \log \Big( \int q(\mathbf{y}_{1:T} \vert \mathbf{y}_0, \mathbf{x}, \mathbf{k}) \frac{p_\theta(\mathbf{y}_{0:T}\vert \mathbf{x}, \mathbf{k})}{q(\mathbf{y}_{1:T} \vert \mathbf{y}_{0}, \mathbf{x}, \mathbf{k})} d\mathbf{y}_{1:T} \Big) \\
&= - \mathbb{E}_{q(\mathbf{y}_0\vert \mathbf{x}, \mathbf{k})} \log \Big( \mathbb{E}_{q(\mathbf{y}_{1:T} \vert \mathbf{y}_0, \mathbf{x}, \mathbf{k})} \frac{p_\theta(\mathbf{y}_{0:T}\vert \mathbf{x}, \mathbf{k})}{q(\mathbf{y}_{1:T} \vert \mathbf{y}_{0}, \mathbf{x}, \mathbf{k})} \Big) \\
&\leq - \mathbb{E}_{q(\mathbf{y}_{0:T}\vert \mathbf{x}, \mathbf{k})} \log \frac{p_\theta(\mathbf{y}_{0:T}\vert \mathbf{x}, \mathbf{k})}{q(\mathbf{y}_{1:T} \vert \mathbf{y}_{0}, \mathbf{x}, \mathbf{k})} \quad \text{ (Jensen's inequality for the concave function)}\\
&= \mathbb{E}_{q(\mathbf{y}_{0:T}\vert \mathbf{x}, \mathbf{k})}\Big[\log \frac{q(\mathbf{y}_{1:T} \vert \mathbf{y}_{0}, \mathbf{x}, \mathbf{k})}{p_\theta(\mathbf{y}_{0:T}\vert \mathbf{x}, \mathbf{k})} \Big] \\
&= \mathbb{E}_{q(\mathbf{y}_{0:T}\vert \mathbf{x}, \mathbf{k})}\Big[\log \frac{q(\mathbf{y}_{1:T} \vert \mathbf{y}_{0})}{p_\theta(\mathbf{y}_{0:T}\vert \mathbf{x}, \mathbf{k})} \Big] = L_\mathrm{VUB}.
\end{split}
\end{equation}

\pagebreak
To convert each term in the equation to be analytically computable, the objective can be further rewritten to be a combination of several KL-divergence and entropy terms:
\begin{equation}
\begin{split}
L_\text{VUB} &= \mathbb{E}_{q(\mathbf{y}_{0:T}\vert \mathbf{x}, \mathbf{k})} \Big[ \log\frac{q(\mathbf{y}_{1:T}\vert\mathbf{y}_0)}{p_\theta(\mathbf{y}_{0:T}\vert \mathbf{x}, \mathbf{k})} \Big] \\
&= \mathbb{E}_q \Big[ \log\frac{\prod_{t=1}^T q(\mathbf{y}_t\vert\mathbf{y}_{t-1})}{ p_\theta(\mathbf{y}_T) \prod_{t=1}^T p_\theta(\mathbf{y}_{t-1} \vert\mathbf{y}_t, \mathbf{x}, \mathbf{k})} \Big] \\
&= \mathbb{E}_q \Big[ -\log p_\theta(\mathbf{y}_T) + \sum_{t=1}^T \log \frac{q(\mathbf{y}_t\vert\mathbf{y}_{t-1})}{p_\theta(\mathbf{y}_{t-1} \vert\mathbf{y}_t, \mathbf{x}, \mathbf{k})} \Big] \\
&= \mathbb{E}_q \Big[ -\log p_\theta(\mathbf{y}_T) + \sum_{t=2}^T \log \frac{q(\mathbf{y}_t\vert\mathbf{y}_{t-1})}{p_\theta(\mathbf{y}_{t-1} \vert\mathbf{y}_t, \mathbf{x}, \mathbf{k})} + \log\frac{q(\mathbf{y}_1 \vert \mathbf{y}_0)}{p_\theta(\mathbf{y}_0 \vert \mathbf{y}_1, \mathbf{x}, \mathbf{k})} \Big] \\
&= \mathbb{E}_q \Big[ -\log p_\theta(\mathbf{y}_T) + \sum_{t=2}^T \log \Big( \frac{q(\mathbf{y}_{t-1} \vert \mathbf{y}_t, \mathbf{y}_0)}{p_\theta(\mathbf{y}_{t-1} \vert\mathbf{y}_t, \mathbf{x}, \mathbf{k})}\cdot \frac{q(\mathbf{y}_t \vert \mathbf{y}_0)}{q(\mathbf{y}_{t-1}\vert\mathbf{y}_0)} \Big) + \log \frac{q(\mathbf{y}_1 \vert \mathbf{y}_0)}{p_\theta(\mathbf{y}_0 \vert \mathbf{y}_1, \mathbf{x}, \mathbf{k})} \Big] \\
&= \mathbb{E}_q \Big[ -\log p_\theta(\mathbf{y}_T) + \sum_{t=2}^T \log \frac{q(\mathbf{y}_{t-1} \vert \mathbf{y}_t, \mathbf{y}_0)}{p_\theta(\mathbf{y}_{t-1} \vert\mathbf{y}_t, \mathbf{x}, \mathbf{k})} + \sum_{t=2}^T \log \frac{q(\mathbf{y}_t \vert \mathbf{y}_0)}{q(\mathbf{y}_{t-1} \vert \mathbf{y}_0)} + \log\frac{q(\mathbf{y}_1 \vert \mathbf{y}_0)}{p_\theta(\mathbf{y}_0 \vert \mathbf{y}_1, \mathbf{x}, \mathbf{k})} \Big] \\
&= \mathbb{E}_q \Big[ -\log p_\theta(\mathbf{y}_T) + \sum_{t=2}^T \log \frac{q(\mathbf{y}_{t-1} \vert \mathbf{y}_t, \mathbf{y}_0)}{p_\theta(\mathbf{y}_{t-1} \vert\mathbf{y}_t, \mathbf{x}, \mathbf{k})} + \log\frac{q(\mathbf{y}_T \vert \mathbf{y}_0)}{q(\mathbf{y}_1 \vert \mathbf{y}_0)} + \log \frac{q(\mathbf{y}_1 \vert \mathbf{y}_0)}{p_\theta(\mathbf{y}_0 \vert \mathbf{y}_1, \mathbf{x}, \mathbf{k})} \Big]\\
&= \mathbb{E}_q \Big[ \log\frac{q(\mathbf{y}_T \vert \mathbf{y}_0)}{p_\theta(\mathbf{y}_T)} + \sum_{t=2}^T \log \frac{q(\mathbf{y}_{t-1} \vert \mathbf{y}_t, \mathbf{y}_0)}{p_\theta(\mathbf{y}_{t-1} \vert\mathbf{y}_t, \mathbf{x}, \mathbf{k})} - \log p_\theta(\mathbf{y}_0 \vert \mathbf{y}_1, \mathbf{x}, \mathbf{k}) \Big] \\
&= \mathbb{E}_q [\underbrace{D_\text{KL}(q(\mathbf{y}_T \vert \mathbf{y}_0) \parallel p_\theta(\mathbf{y}_T))}_{L_T} + \sum_{t=1}^{T-1} \underbrace{D_\text{KL}(q(\mathbf{y}_{t} \vert \mathbf{y}_{t+1}, \mathbf{y}_0) \parallel p_\theta(\mathbf{y}_{t} \vert\mathbf{y}_{t+1}, \mathbf{x}, \mathbf{k}))}_{L_{t}} \underbrace{- \log p_\theta(\mathbf{y}_0 \vert \mathbf{y}_1, \mathbf{x}, \mathbf{k})}_{L_0} ]
\end{split}
\end{equation}

Let’s label each component in the variational upper bound loss separately:
\begin{equation}
\begin{split}
L_\mathrm{VUB} &= L_T + \dots + L_t + \dots + L_0, \\
\text{where } L_T &= D_\text{KL}(q(\mathbf{y}_T \vert \mathbf{y}_0) \parallel p_\theta(\mathbf{y}_T)), \\
L_t &= D_\text{KL}(q(\mathbf{y}_t \vert \mathbf{y}_{t+1}, \mathbf{y}_0) \parallel p_\theta(\mathbf{y}_t \vert\mathbf{y}_{t+1}, \mathbf{x}, \mathbf{k})) \text{ for }1 \leq t \leq T-1, \\
L_0 &= - \log p_\theta(\mathbf{y}_0 \vert \mathbf{y}_1, \mathbf{x}, \mathbf{k}).
\end{split}
\end{equation}

Recall that we need to learn a neural network to approximate the conditioned probability distributions in the reverse diffusion process, 
\begin{equation}
p_{\theta}\left(\mathbf{y}_{t-1} \vert \mathbf{y}_{t}, \mathbf{x},\mathbf{k}\right)=\mathcal{N}\left(\mathbf{y}_{t-1} ; \boldsymbol{\mu}_{\theta}\left(\mathbf{y}_{t}, \mathbf{x}, \mathbf{k}, t\right), \mathbf{\Sigma}_{\theta}\left(\mathbf{y}_{t}, \mathbf{x}, \mathbf{k}, t\right)\right).   
\end{equation}

We aim to train $\boldsymbol{\mu}_\theta$ to predict $\tilde{\boldsymbol{\mu}}_t = \frac{1}{\sqrt{\alpha_t}} \Big( \mathbf{y}_t - \frac{\beta_t}{\sqrt{1 - \bar{\alpha}_t}} \boldsymbol{\epsilon}_t \Big)$. Because $\mathbf{y}_t$ is available as input at training time, we can reparameterize the Gaussian noise term instead to make it predict $\boldsymbol{\epsilon}_t$ from the input $\mathbf{y}_t$ at time step $t$:
\begin{equation}
\boldsymbol{\mu}_\theta(\mathbf{y}_t, \mathbf{x}, \mathbf{k}, t) = \frac{1}{\sqrt{\alpha_t}} \Big( \mathbf{y}_t - \frac{\beta_t}{\sqrt{1 - \bar{\alpha}_t}} \boldsymbol{\epsilon}_\theta(\mathbf{y}_t, \mathbf{x}, \mathbf{k}, t) \Big)
\end{equation}
\begin{equation}
\mathbf{y}_{t-1} = \mathcal{N}(\mathbf{y}_{t-1}; \frac{1}{\sqrt{\alpha_t}} \Big( \mathbf{y}_t - \frac{\beta_t}{\sqrt{1 - \bar{\alpha}_t}} \boldsymbol{\epsilon}_\theta(\mathbf{y}_t, \mathbf{x}, \mathbf{k}, t) \Big), \boldsymbol{\Sigma}_\theta(\mathbf{y}_t, \mathbf{x}, \mathbf{k}, t))
\end{equation}

The loss term $L_t$ is parameterized to minimize the difference between $\boldsymbol{\mu}_\theta$ and $\tilde{\boldsymbol{\mu}}$:
\begin{equation}
\begin{split}
L_t &= D_\mathrm{KL}(q(\mathbf{y}_t \vert \mathbf{y}_{t+1}, \mathbf{y}_0) \parallel p_\theta(\mathbf{y}_t \vert\mathbf{y}_{t+1}, \mathbf{x}, \mathbf{k})) \\
&= \mathbb{E}_{\mathbf{y}_0, \boldsymbol{\epsilon}} \Big[\frac{1}{2 \| \boldsymbol{\Sigma}_\theta(\mathbf{y}_t, \mathbf{x}, \mathbf{k}, t) \|^2_2} \| \tilde{\boldsymbol{\mu}}_t(\mathbf{y}_t, \mathbf{y}_0) - \boldsymbol{\mu}_\theta(\mathbf{y}_t, \mathbf{x}, \mathbf{k}, t)\|^2 \Big] \\
&= \mathbb{E}_{\mathbf{y}_0, \boldsymbol{\epsilon}} \Big[\frac{1}{2  \|\boldsymbol{\Sigma}_\theta \|^2_2} \| \frac{1}{\sqrt{\alpha_t}} \Big( \mathbf{y}_t - \frac{\beta_t}{\sqrt{1 - \bar{\alpha}_t}} \boldsymbol{\epsilon}_t \Big) - \frac{1}{\sqrt{\alpha_t}} \Big( \mathbf{y}_t - \frac{\beta_t}{\sqrt{1 - \bar{\alpha}_t}} \boldsymbol{\epsilon}_\theta(\mathbf{y}_t,\mathbf{x}, \mathbf{k}, t) \Big) \|^2 \Big] \\
&= \mathbb{E}_{\mathbf{y}_0, \boldsymbol{\epsilon}} \Big[\frac{ \beta_t^2 }{2 \alpha_t (1 - \bar{\alpha}_t) \| \boldsymbol{\Sigma}_\theta \|^2_2} \|\boldsymbol{\epsilon}_t - \boldsymbol{\epsilon}_\theta(\mathbf{y}_t,\mathbf{x}, \mathbf{k}, t)\|^2 \Big] \\
&= \mathbb{E}_{\mathbf{y}_0, \boldsymbol{\epsilon}} \Big[\frac{ \beta_t^2 }{2 \alpha_t (1 - \bar{\alpha}_t) \| \boldsymbol{\Sigma}_\theta \|^2_2} \|\boldsymbol{\epsilon}_t - \boldsymbol{\epsilon}_\theta(\sqrt{\bar{\alpha}_t}\mathbf{y}_0 + \sqrt{1 - \bar{\alpha}_t}\boldsymbol{\epsilon}_t, \mathbf{x}, \mathbf{k}, t)\|^2 \Big]. 
\end{split}    
\end{equation} 

Following the standard training process of DPM~\cite{Ho-NIPS2020}, the loss term $L_t$ is parameterized to achieve better training, resulting in a simplified loss:
\begin{equation}
L_t^\mathrm{simple} = \mathbb{E}_{\mathbf{y}_0, \boldsymbol{\epsilon}} \Big[\|\boldsymbol{\epsilon} - \boldsymbol{\epsilon}_\theta(\sqrt{\bar{\alpha}_t}\mathbf{y}_0 + \sqrt{1 - \bar{\alpha}_t}\boldsymbol{\epsilon}, \mathbf{x}, \mathbf{k}, t)\|^2 \Big],
\end{equation}
where $\boldsymbol{\epsilon} \sim \mathcal{N}(\mathbf{0}, \mathbf{I})$.

\subsection{Classifier-free Guided Mask Sampling}
Inspired by the derivation of score-based models~\cite{Song-NIPS2019}, the distribution $p(\mathbf{y}_t|\mathbf{x}, \mathbf{k}, \mathbf{c})$ has the score $\nabla\log p(\mathbf{y}_t|\mathbf{x}, \mathbf{k}, \mathbf{c})$, where $\mathbf{c}$ is a one-hot vector indicating the category of the instance object, which is naturally achieved by classification branch. Via Bayes' rule, we derive the following equivalent form:
\begin{equation}
\begin{aligned}
\nabla \log p\left(\mathbf{y}_t|\mathbf{x},\mathbf{k}, \mathbf{c}\right) &=\nabla \log \left(\frac{p\left(\mathbf{y}_t| \mathbf{x},\mathbf{k}\right) p\left(\mathbf{c} | \mathbf{y}_t,  \mathbf{x},\mathbf{k}\right)}{p(\mathbf{c} | \mathbf{x},\mathbf{k})}\right) \\
&= \nabla\log p\left(\mathbf{y}_t| \mathbf{x},\mathbf{k}\right) + \nabla\log p\left(\mathbf{c} | \mathbf{y}_t,  \mathbf{x},\mathbf{k}\right) - \nabla\log p(\mathbf{c} | \mathbf{x},\mathbf{k}) \\
&=\underbrace{\nabla \log p\left(\mathbf{y}_t| \mathbf{x},\mathbf{k}\right)}_{\text {guidance-agnostic score }}+\underbrace{\nabla \log p\left(\mathbf{c} | \mathbf{y}_t, \mathbf{x},\mathbf{k}\right)}_{\text {adversarial gradient}}.
\end{aligned} \label{eq:cls_guide}
\end{equation}

Classifier guidance~\cite{Dhariwal-NIPS2021} adjusts the adversarial gradient of the noisy classifier by a $\omega$ hyper-parameter term to introduce fine-grained control to either encourage or dissuade the model from accepting the conditioning information. The score function is thus described as follows:
\begin{equation}
\nabla \log p\left(\mathbf{y}_t|\mathbf{x},\mathbf{k}, \mathbf{c}\right)
=\nabla \log p\left(\mathbf{y}_t| \mathbf{x},\mathbf{k}\right)+\omega\nabla \log p\left(\mathbf{c} | \mathbf{y}_t, \mathbf{x},\mathbf{k}\right). \label{eq:cls_gamma}
\end{equation}

Inspired by classifier-free guidance~\cite{Ho-NIPSW2021}, 
we arrange the Eq.~\ref{eq:cls_guide} to obtain the adversarial gradient term: 
\begin{equation}
\nabla \log p\left(\mathbf{c} | \mathbf{y}_t, \mathbf{x},\mathbf{k}\right) = \nabla \log p\left(\mathbf{y}_t|\mathbf{x},\mathbf{k}, \mathbf{c}\right) - \nabla \log p\left(\mathbf{y}_t| \mathbf{x},\mathbf{k}\right).     
\end{equation}

After that, we substitute the adversarial gradient term above into Eq.~\ref{eq:cls_gamma}, resulting:
\begin{equation}
\nabla \log p\left(\mathbf{y}_t|\mathbf{x},\mathbf{k}, \mathbf{c}\right) = \underbrace{\omega\nabla \log p\left(\mathbf{y}_t| \mathbf{x},\mathbf{k}, \mathbf{c}\right)}_{\text {guidance-specific score }}+\underbrace{(1-\omega)\nabla \log p\left(\mathbf{y}_t|\mathbf{x},\mathbf{k}\right)}_{\text {guidance-agnostic score}}.
\end{equation}

We can formulate the derivative of the logarithm of its density function as follows:
\begin{equation}
\resizebox{0.9\hsize}{!}{$
\begin{split}
\nabla_{\mathbf{y}_t} \log p\left(\mathbf{y}_t|\mathbf{x},\mathbf{k}, \mathbf{c}\right) &= \mathbb{E}_{\mathbf{y}_{t+1}}[\nabla_{\mathbf{y}_t} \log p\left(\mathbf{y}_t|\mathbf{y}_{t+1},\mathbf{x},\mathbf{k}, \mathbf{c}\right)] = \mathbb{E}_{\mathbf{y}_{t+1}}\left[\nabla_{\mathbf{y}_t} -\frac{1}{2}\left(\frac{\mathbf{y}_t - \boldsymbol{\mu}_\theta}{\boldsymbol{\sigma}_\theta}\right)^2\right] = \mathbb{E}_{\mathbf{y}_{t+1}}\left[-\frac{\mathbf{y}_t-\boldsymbol{\mu}_\theta}{\boldsymbol{\sigma}_\theta^2}\right] \\ 
&= \mathbb{E}_{\mathbf{y}_{t+1}}\left[-\frac{\boldsymbol{\epsilon}_\theta}{\boldsymbol{\sigma}_\theta}\right] = \mathbb{E}_{\mathbf{y}_{t+1}}\left[-\frac{\boldsymbol{\epsilon}_\theta(\mathbf{y}_t, \mathbf{x}, \mathbf{k}, \mathbf{c}, t)}{\sqrt{1-\bar{\alpha}_t}}\right] = -\frac{1}{\sqrt{1-\bar{\alpha}_t}}\boldsymbol{\epsilon}_\theta(\mathbf{y}_t, \mathbf{x}, \mathbf{k}, \mathbf{c}, t).
\end{split}$}
\end{equation}

As $\nabla \log p\left(\mathbf{y}_t|\mathbf{x},\mathbf{k}, \mathbf{c}\right) = -\frac{1}{\sqrt{1-\bar{\alpha}_t}}\boldsymbol{\epsilon}_\theta(\mathbf{y}_t, \mathbf{x}, \mathbf{k}, \mathbf{c}, t)$, we have the equivalent form:
\begin{equation}
\boldsymbol{\hat{\epsilon}}_\theta(\mathbf{y}_t, \mathbf{x}, \mathbf{k}, \mathbf{c}, t) = \underbrace{\omega\boldsymbol{\epsilon}_\theta(\mathbf{y}_t, \mathbf{x}, \mathbf{k}, \mathbf{c}, t)}_{\text {guidance-specific score }}+\underbrace{(1-\omega)\boldsymbol{\epsilon}_\theta(\mathbf{y}_t, \mathbf{x}, \mathbf{k}, t)}_{\text {guidance-agnostic score}}.
\end{equation}

\section{Comprehensive Experiments}
\setlabel{Comprehensive Experiments}{sec:experiments}
\subsection{Comparison with State-of-the-art Methods}
\textbf{Results on both COCO base and novel classes (COCO-All).} In this experiment, we aim to predict all $80$ COCO classes and report the standard evaluation metrics, including mAP and $AP_50$. Following the newly introduced evaluation process of Ganea~\etal~\shortcite{Ganea-CVPR2021}, we report the performance of the base and novel classes independently as well as all classes. We compared our MaskDiff against state-of-the-art methods of FSOD (\eg, ONCE~\cite{Perez-CVPR2020}, TFA~\cite{Wang-ICML2020}, FSDetView~\cite{Xiao-ECCV2020}, and LEAST~\cite{Li-2021}) and FSIS (\eg, Meta-RCNN~\cite{Yan-ICCV2019}, such as MTFA~\cite{Ganea-CVPR2021}, iMTFA~\cite{Ganea-CVPR2021}, iFS-RCNN~\cite{Nguyen-CVPR2022}, and a fully-converged Mask R-CNN model fine-tuned on the novel classes (MRCN+ft-full)~\cite{He-ICCV2017}). Results in Table~\ref{table:base_novel} show that MaskDiff consistently outperforms state-of-the-art methods on both FSOD and FSIS tasks on both base and novel classes for all numbers of provided shots. This implies MaskDiff has ability to adapt to novel classes while embracing performance in base classes. 

\begin{table*}[t!]
\centering

\resizebox{\linewidth}{!}{
\begin{tabular}{c|l|c|cc|cc|cc|cc|cc|cc}
\toprule
\multirow{3}{*}{\textbf{Shots}} &  \multirow{3}{*}{\textbf{Method}} & \multirow{3}{*}{\textbf{Published}} & \multicolumn{6}{c|}{\textbf{Object Detection}} & \multicolumn{6}{c}{\textbf{Instance Segmentation}} \\ \cmidrule{4-15}
 & & & \multicolumn{2}{c|}{\textbf{All Classes}} & \multicolumn{2}{c|}{\textbf{Base Classes}} & \multicolumn{2}{c|}{\textbf{Novel Classes}} & \multicolumn{2}{c|}{\textbf{All Classes}} & \multicolumn{2}{c|}{\textbf{Base Classes}} & \multicolumn{2}{c}{\textbf{Novel Classes}} \\  
 & & & 
  \textbf{AP} &
  \textbf{AP$_{50}$} &
  \textbf{AP} & 
  \textbf{AP$_{50}$} &  
  \textbf{AP} &
  \textbf{AP$_{50}$} &
  \textbf{AP} &
  \textbf{AP$_{50}$} &
  \textbf{AP} &
  \textbf{AP$_{50}$} &
  \textbf{AP} &
  \textbf{AP$_{50}$} \\
\midrule
\multirow{5}{*}{1} 
& ONCE~\cite{Perez-CVPR2020} & CVPR 2020 & 13.60 & - & 17.90  & -  & 0.70  & - & - & - & - & - & - & - \\
 & TFA~\cite{Wang-ICML2020} & ICML 2020 & 24.40 & 39.80 & 31.90 &  51.80 & 1.90  & 3.80 & - & - & - & - & - & - \\
 & FSDetView~\cite{Xiao-ECCV2020} & ECCV 2020 & 20.15  & 29.87 & 25.75 & 39.86 & 3.35  & 6.11  & - & - & - & - & - & - \\
 & LEAST~\cite{Li-2021} & arXiv 2021 & 7.50 & -  & 24.60  &  - & 4.40  &  - & - & - & - & - & - & - \\
\cmidrule{2-15}
 & MRCN+ft-full~\cite{He-ICCV2017} & ICCV 2017 & 10.21 & 21.58 & 17.63 & 26.32 & 0.74 & 2.33 & 9.88 & 19.25 & 15.57 & 24.18 & 0.64 & 2.14 \\
 & MTFA~\cite{Ganea-CVPR2021} & CVPR 2021  & 24.32 & 39.64 & 31.73 & 51.49 & 2.10 & 4.07 & 22.98 & 37.48 & 29.85 & 48.64 & 2.34 & 3.99\\
 & iMTFA~\cite{Ganea-CVPR2021} & CVPR 2021  & 21.67 & 31.55 & 27.81 & 40.11 & 3.23 & 5.89 & 20.13 & 30.64 & 25.9 & 39.28 & 2.81 & 4.72\\
 & iFS-RCNN~\cite{Nguyen-CVPR2022} & CVPR 2022 & 31.19 & 52.83 & 40.08 & 71.14 & 4.54 & 10.29 & 28.45 & 46.72 & 36.35 & 63.11 & 3.95 & 7.89 \\
\rowcolor{lightgray} & \textbf{MaskDiff (Ours)} & - &  \textbf{32.59} & \textbf{54.61} & \textbf{41.23} & \textbf{72.85} & \textbf{5.26} & \textbf{11.35} & \textbf{29.42} & \textbf{48.37} & \textbf{37.59} & \textbf{64.26} & \textbf{4.85} & \textbf{8.73}\\
\midrule
\multirow{5}{*}{5} 
& ONCE~\cite{Perez-CVPR2020} & CVPR 2020 & 13.70 & - & 17.90 & - & 1.00 & - & - & - & - & - & - & - \\
 & TFA~\cite{Wang-ICML2020} & ICML 2020 & 25.90  & 41.20  & 32.30  & 50.50  & 7.00 & 13.30 & - & - & - & - & - & - \\
 & FSDetView~\cite{Xiao-ECCV2020} & ECCV 2020 & 20.92 & 30.23 & 25.05 & 40.32 & 8.53 & 15.79  & - & - & - & - & - & - \\
 & LEAST~\cite{Li-2021} & arXiv 2021 & 13.70  &  -  &  25.20 &  - & 9.40 & -  & - & - & - & - & - & - \\
\cmidrule{2-15}
 & MRCN+ft-full~\cite{He-ICCV2017} & ICCV 2017 &  12.31 & 23.69 & 19.43 & 28.12 & 1.15 & 2.09 & 11.28 & 22.36 & 17.89 & 26.78 & 1.17 & 2.58 \\
 & MTFA~\cite{Ganea-CVPR2021} & CVPR 2021 &  26.39 & 41.52 & 33.11 & 51.49 & 6.22 & 11.63 & 25.07 & 39.95 & 31.29 & 49.55 & 6.38 & 11.14\\
 & iMTFA~\cite{Ganea-CVPR2021} & CVPR 2021 &  19.62 & 28.06 & 24.13 & 33.69 & 6.07 & 11.15 & 18.22 & 27.10 & 22.56 & 33.25 & 5.19 & 8.65\\
 & iFS-RCNN~\cite{Nguyen-CVPR2022} & CVPR 2022 &  32.52 & 54.30 & 40.06 & 71.19 & 9.91 & 19.24 & 29.89 & 48.22 & 36.33 & 62.81 & 8.80 & 15.73 \\
\rowcolor{lightgray} & \textbf{MaskDiff (Ours)} & - &  \textbf{33.45} & \textbf{56.19} & \textbf{41.51} & \textbf{72.28} & \textbf{10.49} & \textbf{21.07} & \textbf{30.48} & \textbf{50.35} & \textbf{38.12} & \textbf{64.36} & \textbf{9.43} & \textbf{17.12} \\
\midrule
\multirow{5}{*}{10} 
& ONCE~\cite{Perez-CVPR2020} & CVPR 2020 & 13.70 & - & 17.90  & - & 1.20 & - & - & - & - & - & - & - \\
 & TFA~\cite{Wang-ICML2020} & ICML 2020 & 26.60 & 42.20 & 32.40 &  50.60 &  9.10 & 17.10  & - & - & - & - & - & - \\
 & FSDetView~\cite{Xiao-ECCV2020} & ECCV 2020 & 21.74 & 31.55  & 24.82 & 40.26 &  12.50 & 26.09  & - & - & - & - & - & - \\
 & LEAST~\cite{Li-2021} & arXiv 2021 & 16.20  & - & 23.10  & -  &  12.50 & -  & - & - & - & - & - & - \\
\cmidrule{2-15}
 & MRCN+ft-full~]\cite{He-ICCV2017} & ICCV 2017 &  12.44 & 24.39 & 20.57 & 29.72 & 2.33 & 5.64 & 12.15 & 23.29 & 18.08 & 27.53 & 1.86 & 4.25 \\
 & MTFA~\cite{Ganea-CVPR2021} & CVPR 2021 &  27.44 & 42.84 & 33.83 & 52.04 & 8.28 & 15.25 & 25.97 & 41.28 & 31.84 & 50.17 & 8.36 & 14.58 \\
 & iMTFA~\cite{Ganea-CVPR2021} & CVPR 2021 &  19.26 & 27.49 & 23.36 & 32.41 & 6.97 & 12.72 & 17.87 & 26.46 & 21.87 & 32.01 & 5.88 & 9.81\\
 & iFS-RCNN~\cite{Nguyen-CVPR2022} & CVPR 2022 &  33.02 & 56.15 & 40.05 & 69.84 & 12.55 & 25.97 & 30.41 & 49.54 & 36.32 & 63.29 & 10.06 & 19.72\\
\rowcolor{lightgray} & \textbf{MaskDiff (Ours)} & - &  \textbf{35.21} & \textbf{59.80} & \textbf{42.17} & \textbf{72.36} & \textbf{14.04} & \textbf{28.33} & \textbf{31.89} & \textbf{52.15} & \textbf{38.55} & \textbf{66.48} & \textbf{11.84} & \textbf{21.27} \\
\bottomrule
\end{tabular}
}
\caption{FSOD and FSIS performance on COCO dataset for both base and novel classes (COCO-All). MaskDiff outperforms state-of-the-art methods. The best performance is marked in boldface. }
\label{table:base_novel}
\vspace{-3mm}
\end{table*}

\begin{table}[!t]
\centering

\resizebox{\linewidth}{!}{
\begin{tabular}{c|l|cccc|cccc}
\toprule
\multirow{2}{*}{\textbf{Shots}} & \multirow{2}{*}{\textbf{Method}} & \multicolumn{4}{c|}{\textbf{Detection}} & \multicolumn{4}{c}{\textbf{Segmentation}} \\ \cmidrule{3-10}
 &
 &
  \textbf{AP} &
  \textbf{AP$_{50}$} &
  \textbf{AP$_{75}$} &
  \textbf{AP$_{95}$} &
  \textbf{AP} & 
  \textbf{AP$_{50}$} &  
  \textbf{AP$_{75}$} & 
  \textbf{AP$_{95}$} \\
\midrule
\multirow{5}{*}{1} & MRCN+ft-full~\cite{He-ICCV2017} & 0.86 & 2.58 & 1.29 & 0.25 & 0.75 & 2.47 & 1.13 & 0.19 \\
& Meta-RCNN~\cite{Yan-ICCV2019} & - & - & - & - & - & - & - & - \\
 & MTFA~\cite{Ganea-CVPR2021} & 2.49 & 4.86 & 2.26 & 0.89 & 2.67 & 4.58 & 2.77 & 0.77 \\
 & iMTFA~\cite{Ganea-CVPR2021} & 3.31 & 6.05 & 3.15 & 1.02 & 2.85 & 4.76 & 2.90 & 0.86\\
 & iFS-RCNN~\cite{Nguyen-CVPR2022} & 6.37 & 12.90 & 3.42 & 1.76 & 5.56 & 11.24 & 3.19 & 1.54\\
\rowcolor{lightgray} & \textbf{MaskDiff (Ours)} & \textbf{7.13} & \textbf{13.51} & \textbf{4.57} & \textbf{3.79} & \textbf{6.26} & \textbf{12.45} & \textbf{4.41} & \textbf{3.60} \\
\midrule
\multirow{5}{*}{5} & MRCN+ft-full~\cite{He-ICCV2017} & 1.32 & 3.10 & 1.13 & 0.46 & 1.31 & 2.69 & 1.12 & 0.39 \\
& Meta-RCNN~\cite{Yan-ICCV2019} & 3.50 & 9.90 & 1.20 & - & 2.80 & 6.90 & 1.70 & - \\
 & MTFA~\cite{Ganea-CVPR2021} & 6.61 & 12.32 & 6.39 & 1.24 & 6.63 & 11.59 & 6.67 & 1.34 \\
 & iMTFA~\cite{Ganea-CVPR2021} & 6.21 & 11.26 & 3.03 & 1.17 & 5.20 & 8.76 & 2.77 & 1.43  \\
 & iFS-RCNN~\cite{Nguyen-CVPR2022} & 10.54 & 21.07 & 6.11 & 4.39 & 9.44 & 19.15 & 5.86 & 3.41\\
\rowcolor{lightgray} & \textbf{MaskDiff (Ours)} & \textbf{11.72} & \textbf{22.48} & \textbf{8.47} & \textbf{7.36} & \textbf{10.09} & \textbf{20.64} & \textbf{7.71} & \textbf{5.92} \\
\midrule
\multirow{5}{*}{10} & MRCN+ft-full~\cite{He-ICCV2017} & 2.52 & 5.78 & 1.91 & 0.97 & 1.93 & 4.68 & 1.32 & 0.84 \\
& Meta-RCNN~\cite{Yan-ICCV2019} & 5.60 & 14.20 & 3.00 & - & 4.40 & 10.60 & 3.30 & - \\
 & MTFA~\cite{Ganea-CVPR2021} & 8.52 & 15.53 & 8.44 & 3.37 & 8.40 & 14.62 & 8.46 & 3.15 \\
 & iMTFA~\cite{Ganea-CVPR2021} & 7.14 & 12.91 & 6.93 & 3.19 & 5.94 & 9.96 & 6.09 & 3.04 \\
 & iFS-RCNN~\cite{Nguyen-CVPR2022} & 11.27 & 22.15 & 8.12 & 5.46 & 10.22 & 20.61 & 7.39 & 4.60 \\
\rowcolor{lightgray} & \textbf{MaskDiff (Ours)} & \textbf{12.18} & \textbf{23.61} & \textbf{10.34} & \textbf{8.12} & \textbf{11.01} & \textbf{21.58} & \textbf{9.27} & \textbf{6.43} \\
\bottomrule
\end{tabular}
}
\caption{FSOD and FSIS performance on only COCO novel classes (COCO-Novel). MaskDiff outperforms state-of-the-art methods. 
There is no report of Meta-RCNN~\cite{Yan-ICCV2019} on one-shot.}
\label{table:novel_full}
\end{table}

\textbf{Results on only COCO novel classes (COCO-Novel).} Table~\ref{table:novel_full} reports our results on only COCO novel classes, in which we compared MaskDiff against state-of-the-art FSIS methods (\eg, Meta-RCNN~\cite{Yan-ICCV2019}, MTFA~\cite{Ganea-CVPR2021}, iMTFA~\cite{Ganea-CVPR2021}, iFS-RCNN~\cite{Nguyen-CVPR2022}, and MRCN+ft-full~\cite{He-ICCV2017}). Experiments demonstrated the superior performance of MaskDiff on both detection and segmentation tasks consistently on all $K-$shots. MaskDiff surpasses the recent iFS-RCNN~\cite{Nguyen-CVPR2022} and drastically outperforms the others by a large margin in terms of AP. This indicates that our proposal is better than previous strategies including episodic-training (Meta-RCNN~\shortcite{Yan-ICCV2019}), class-specific mask predictor (MTFA~\shortcite{Ganea-CVPR2021}) and class-agnostic one (iMTFA~\shortcite{Ganea-CVPR2021}, iFS-RCNN~\shortcite{Nguyen-CVPR2022}).

\begin{figure*}[!t]
    \centering
    \includegraphics[width=\textwidth]{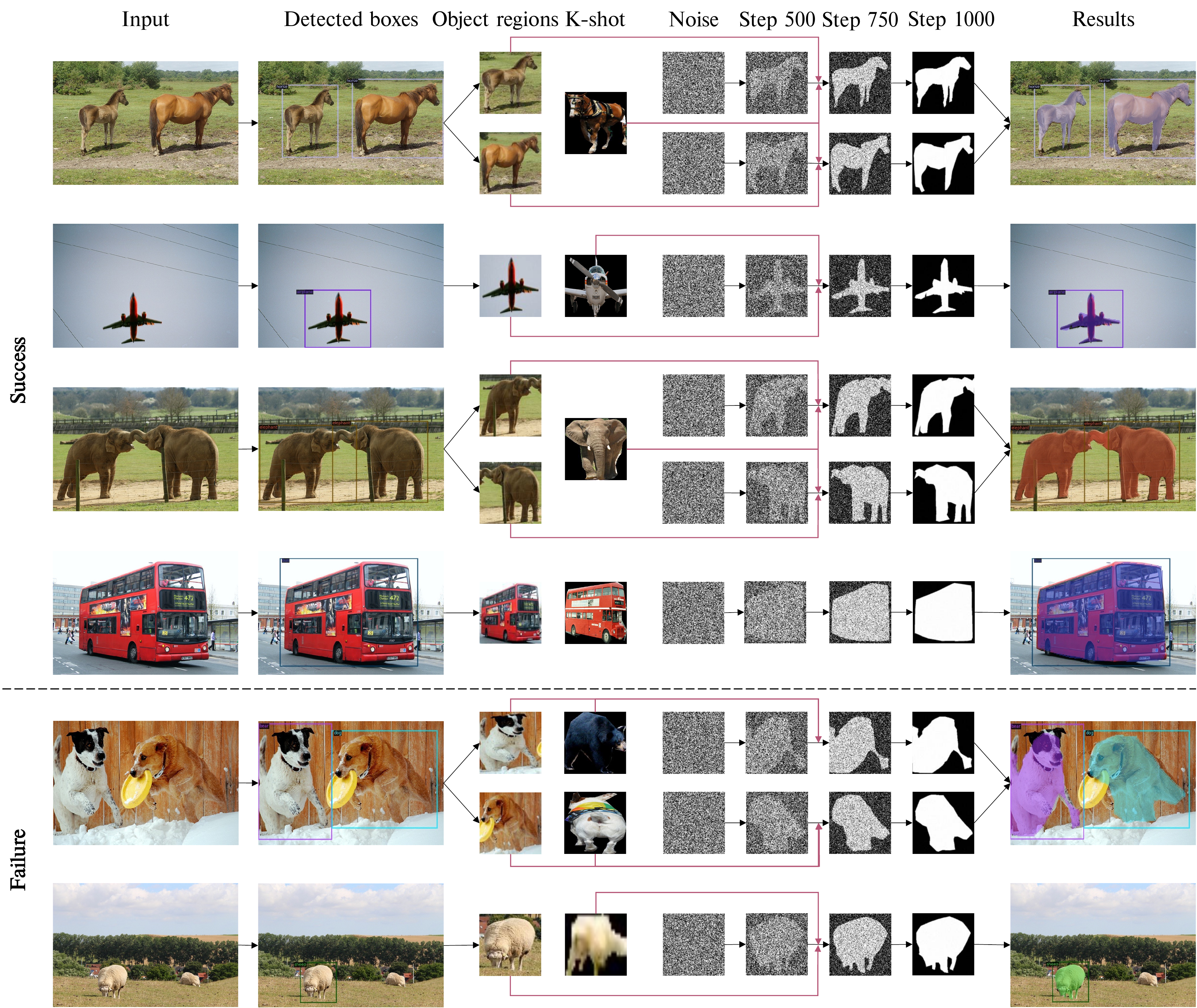}
    \caption{Inference examples. Successful (top four rows) and failure cases (bottom two rows) when training and inference on the one-shot setting for the COCO novel classes. Failures include wrong classification ($5$th row), miss detection, and imprecise instance segmentation (the bottom row).}
    \label{fig:qualitative_full}
    \vspace{-3mm}
\end{figure*}

\begin{table}[t!]
\centering
\resizebox{0.8\linewidth}{!}{
\begin{tabular}{l|cc|c}
\midrule
 \multirow{2}{*}{\textbf{Method}} & \multicolumn{2}{c|}{\textbf{Training Time (Hours)}} & \multirow{2}{*}{\textbf{Inference Time (FPS)}} \\ 
 &
  \textbf{Base} &
  \textbf{Fine-tuning} & 
 \\
\midrule
 MRCN+ft-full~\cite{He-ICCV2017} & 24.6 & 12.8 & 10.38 \\
 MTFA~\cite{Ganea-CVPR2021} & 24.8 & 12.7  & 10.32 \\
 iMTFA~\cite{Ganea-CVPR2021} & \textbf{24.2} & \textbf{12.4} & \textbf{11.75} \\
 iFS-RCNN~\cite{Nguyen-CVPR2022} & 28.5 & 15.6 & 10.28 \\
 \textbf{MaskDiff (Ours)} & 32.3 & 16.1 & 8.27\\
\midrule
\end{tabular}}
\vspace{-3mm}
\caption{Runtime evaluation of FSIS performance on COCO-All dataset. The best performance is marked in boldface.}
\label{table:runtime}
\vspace{-4mm}
\end{table}

\textbf{Additional comparison}. Our work uses a different training and evaluation protocol than FAPIS~\cite{Nguyen-CVPR2021} and DTN~\cite{Wang-MM2022}, two SOTA methods in FSIS. Unlike these works, we follow TFA’s training and evaluation strategy~\cite{Wang-ICML2020}, which is more standard and common. For a fair comparison, we additionally compared our MaskDiff using the same evaluation protocol as these works. As seen in Table~\ref{table:additional_experiments}, our MaskDiff significantly outperforms these two methods by a large margin.
\begin{table}[!t]
\footnotesize
\centering
\resizebox{0.7\linewidth}{!}{
\begin{tabular}{l|cc|cc}
\toprule
 \multirow{2}{*}{\textbf{Method}} & \multicolumn{2}{c|}{\textbf{Detection}} & \multicolumn{2}{c}{\textbf{Segmentation}} \\ 
 &
  \textbf{\#1} &
  \textbf{\#5} &
  \textbf{\#1} & 
  \textbf{\#5}  \\
\midrule
FAPIS~\cite{Nguyen-CVPR2021} & 21.20 & 23.60  & 19.00 & 21.20  \\
DTN~\cite{Wang-MM2022} & 23.80 & 25.30 & 21.50 & 23.40 \\
\rowcolor{lightgray}  \textbf{MaskDiff (Ours)} & \textbf{26.48} & \textbf{28.17} & \textbf{24.73} & \textbf{26.25} \\
\bottomrule
\end{tabular}
}
\caption{Additional experiments on COCO-$20^i$ dataset ($i$ denotes the number of shots). We use the same training and evaluation protocol of compared methods~\cite{Nguyen-CVPR2021, Wang-MM2022} for a fair evaluation.}
\label{table:additional_experiments}
\vspace{-3mm}
\end{table}

\textbf{Qualitative evaluation.} We trained and performed inference of MaskDiff on one-shot setting ($K=1$) on the COCO novel classes. The results are demonstrated in Fig.~\ref{fig:qualitative_full}. The top four rows illustrate successful cases, while failure cases are shown in the bottom four row. Regarding the failure ones, a white ``dog'' is classified as ``bear'' and imperfectly segmented. The other dog is also imprecisely delineated despite being correctly categorized (the 5th row). Miss detection are shown in the last two rows. Imprecise segmentation is illustrated in the last three rows.
Apparently, misclassification and inadequate information from training samples lead to poor segmentation results.
\begin{table}[!t]
\centering
\resizebox{0.95\linewidth}{!}{
\begin{tabular}{c|cc|cccc|cccc}
\toprule
\multirow{2}{*}{\textbf{Shots}} & \multicolumn{2}{c|}{\textbf{MaskDiff}} & \multicolumn{4}{c|}{\textbf{Detection}} & \multicolumn{4}{c}{\textbf{Segmentation}} \\ \cmidrule{2-11}
 &
 Guidance & Two-Stage
 &
  \textbf{AP} &
  \textbf{AP$_{50}$} &
  \textbf{AP$_{75}$} &
  \textbf{AP$_{95}$} &
  \textbf{AP} & 
  \textbf{AP$_{50}$} &  
  \textbf{AP$_{75}$} & 
  \textbf{AP$_{95}$} \\
\midrule
\rowcolor{lightgray}
\multirow{4}{*}{1} 
 & \ding{51} & \ding{51} & 7.13 & 13.51 & 4.57 & 3.79 & 6.26 & 12.45 & 4.41 & 3.60\\
 & \ding{55} & \ding{51} & 6.89 & 13.42 & 4.31 & 3.47 & 6.07 & 12.33 & 4.19 & 3.28\\
 & \ding{51} & \ding{55} & 5.85 & 11.04 & 2.67 & 1.09 & 5.13 & 10.82 & 2.40 & 0.97 \\
 & \ding{55} & \ding{55} & 5.56 & 10.83 & 2.34 & 0.81 & 4.92 & 10.73 & 2.25 & 0.70 \\
\midrule
\rowcolor{lightgray}
\multirow{4}{*}{5} 
 & \ding{51} & \ding{51} & 11.72 & 22.48 & 8.47 & 7.36 & 10.09 & 20.64 & 7.71 & 5.92 \\
 & \ding{55} & \ding{51} & 11.45 & 22.26 & 8.12 & 6.97 & 9.74 & 20.52 & 7.43 & 5.24 \\
 & \ding{51} & \ding{55} & 10.13 & 20.95 & 5.79 & 3.84 & 8.91 & 18.14 & 5.32 & 3.70 \\
 & \ding{55} & \ding{55} & 9.88 & 20.84 & 5.54 & 3.61 & 8.76 & 17.98 & 5.16 & 3.42 \\
\midrule
\rowcolor{lightgray}
\multirow{4}{*}{10}
 & \ding{51} & \ding{51} & 12.18 & 23.61 & 10.34 & 8.12 & 11.01 & 21.58 & 9.27 & 6.43 \\
 & \ding{55} & \ding{51} & 12.05 & 23.32 & 10.17 & 7.98 & 10.76 & 20.63 & 9.04 & 6.28 \\
 & \ding{51} & \ding{55} & 10.86 & 21.55 & 6.39 & 4.41 & 10.02 & 19.48 & 7.13 & 4.37 \\
 & \ding{55} & \ding{55} & 10.59 & 20.96 & 6.14 & 4.25 & 9.68 & 18.83 & 6.90 & 4.06 \\
\bottomrule
\end{tabular}
}
\caption{Our ablation study on COCO-Novel dataset. MaskDiff with classifier-free guided mask sampling and diffusion based two-stage training strategy achieves the best performance. Removing the guidance, the effectiveness of the network witnesses a gradual decrease, while not using the proposed training leads to a drastic drop in AP.}
\vspace{-4mm}
\label{table:ablation}
\end{table}

\textbf{Runtime evaluation.} We evaluate the runtime of FSIS methods by training on the COCO-All train set and inference on the test set and measure the training time and frame per second (FPS) of each method on different $K-$shot settings ($K=1,5,10)$, respectively. Table~\ref{table:runtime} illustrates that our MaskDiff achieves $8.27 \pm 0.09$ FPS with $K-$shot settings, which is acceptable for the FSIS task. Although other methods can run faster than our method, which is easily explained that the diffusion process requires more time to perform than conventional CNN approaches, we remark that for few-shot learning tasks, effectiveness (\ie, mAP) should be prioritized rather than efficiency (\ie, processing time). Currently, all methods have not achieved real-time performance.

\subsection{Ablation Study}
\textbf{Effectiveness of two-stage training strategy.} To demonstrate the effect of the proposed two-stage training strategy on FSIS performance, we jointly trained both object detector heads and our mask distribution modeling head in a single stage. In other words, we remove the first stage and utilize only the second stage. In the early steps, the object detector cannot localize and classify the objects precisely. Given unsatisfactory localized object regions, the mask distribution modeling head is much more difficult to train, leading to generate not good segmentation masks. Table~\ref{table:ablation} indicates a considerable collapse in FSIS results. 

\begin{figure}[!t]
    \centering
    \includegraphics[width=0.7\linewidth]{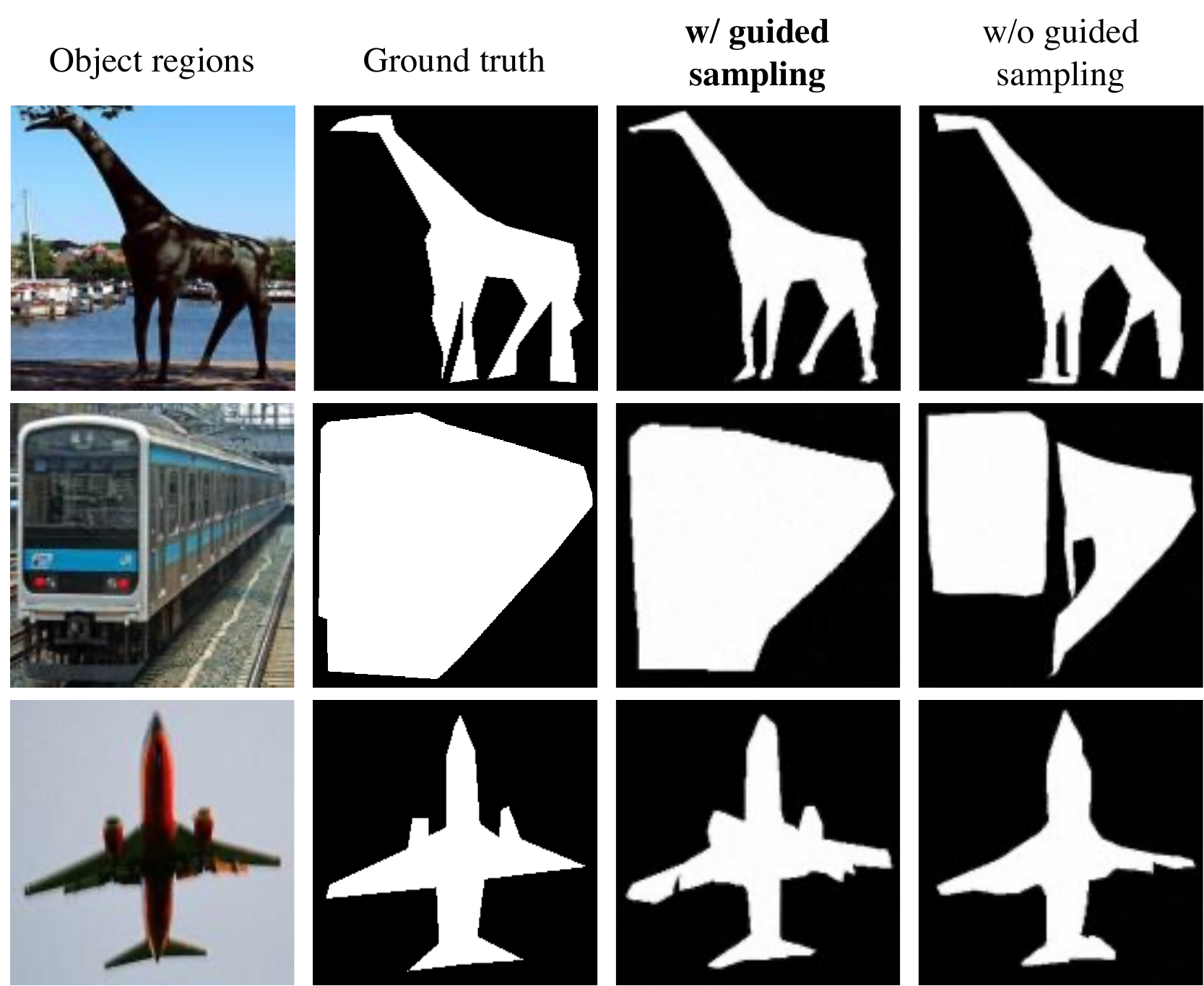}
    \caption{Qualitative results of inference procedure of diffusion model with and without guided sampling. It is apparent that guided sampling can generate more organized and semantic content.}    
    \label{fig:guided_sampling_ex}
    \vspace{-3mm}
\end{figure}

\textbf{Significance of classifier-free guided mask sampling.} 
We studied the effectiveness as well as stability of classifier-free guided mask sampling by comparing the performance MaskDiff with and without classifier-free guided mask sampling. From Table~\ref{table:ablation}, we conclude that classifier-free guided mask sampling boosts the performance of FSIS. Furthermore, we evaluated the standard deviation of testing results on 10 different random seeds when running on a specific sample of training shots. In Table~\ref{table:significance_cls}, we can see that the standard deviation of MaskDiff with guidance over 10 different random seeds is less than that of without guidance. In other words, the classifier-free guided mask sampling makes the mask distribution modeling head more stable. Figure~\ref{fig:guided_sampling_ex} demonstrates the qualitative results of classifier-free guided mask sampling. Binary masks with classifier-free guided mask sampling display more semantic information and produce more organized content.

\begin{table}[!t]
\centering
\resizebox{0.6\columnwidth}{!}{
\begin{tabular}{c|c|cc}
\toprule
\multirow{2}{*}{\textbf{Shots}} & \multirow{2}{*}{\textbf{MaskDiff}} &
\multicolumn{2}{c}{\textbf{Segmentation}} \\ 
 &
 &
  \textbf{AP} & 
  \textbf{AP$_{50}$} \\
\midrule
\multirow{2}{*}{1} 
 &  w/o guided sampling  & 6.04 $\pm$ 0.09 & 12.29 $\pm$ 0.12 \\
 &  \textbf{w/ guided sampling}  & 6.23 $\pm$ \textbf{0.07} & 12.47 $\pm$ \textbf{0.08} \\
\midrule
\multirow{2}{*}{5} 
 &  w/o guided sampling & 9.71 $\pm$ 0.06  & 20.51 $\pm$ 0.10 \\
 &  \textbf{w/ guided sampling} & 10.14 $\pm$ \textbf{0.05} & 20.65 $\pm$ \textbf{0.06} \\
\midrule
\multirow{2}{*}{10}
 &  w/o guided sampling & 10.77 $\pm$ 0.06 & 20.68 $\pm$ 0.09\\
 &  \textbf{w/ guided sampling} & 11.04 $\pm$ \textbf{0.04} & 21.60 $\pm$ \textbf{0.06} \\
\bottomrule
\end{tabular}
}
\caption{We train our MaskDiff on a specific set of training samples and perform inference on 10 different random seeds. Then, we compute the mean and standard deviation (std) AP of those runs. MaskDiff with guided sampling not only outperforms the one without guidance but also is more stable (less std).}
\label{table:significance_cls}
\vspace{-3mm}
\end{table}

\textbf{Stability analysis.} The performance of various methods has been generally evaluated based on the evaluation protocol proposed by Wang~\etal~\shortcite{Wang-ICML2020}. In particular, a certain model is trained on a different set of training samples and is evaluated on the same test set to report the mean result. We further computed the standard deviation of different runs and compared the stability with state-of-the-art methods. Table~\ref{table:stability} shows that our MaskDiff achieves the highest performance and the lowest standard deviation in terms of AP, indicating that MaskDiff is the most stable and robust to different $K$ training samples. 

\begin{table}[t!]
\centering
\resizebox{0.8\linewidth}{!}{
\begin{tabular}{c|l|c|c|c}
\toprule
\textbf{Shots} & \textbf{Method} & \textbf{All Classes} & \textbf{Base Classes} & \textbf{Novel Classes} \\  
\midrule
\multirow{5}{*}{1} & MRCN+ft-full~\cite{He-ICCV2017} & 9.88 $\pm$ 0.31  & 15.57 $\pm$ 0.30 & 0.64 $\pm$ 0.26 \\
 & MTFA~\cite{Ganea-CVPR2021} & 22.98 $\pm$ 0.24 & 29.85 $\pm$ 0.35 & 2.34 $\pm$ 0.31 \\
 & iMTFA~\cite{Ganea-CVPR2021} & 20.13 $\pm$ 0.28 & 25.90 $\pm$ 0.32 & 2.81 $\pm$ 0.37 \\
 & iFS-RCNN~\cite{Nguyen-CVPR2022}  & 28.45 $\pm$ 0.12 & 36.35 $\pm$ 0.01 &  3.95 $\pm$ 0.48\\
\rowcolor{lightgray} & \textbf{MaskDiff (Ours)} & 29.42 \textbf{{$\pm$ 0.09}} & 37.59  \textbf{{$\pm$ 0.0093}} & 4.85 \textbf{{$\pm$ 0.24}}\\
\midrule
\multirow{6}{*}{5} & MRCN+ft-full~\cite{He-ICCV2017} & 11.28 $\pm$ 0.35 & 17.89 $\pm$ 0.46 & 1.17 $\pm$ 0.22\\
 & MTFA~\cite{Ganea-CVPR2021} & 25.07 $\pm$ 0.17 & 31.29 $\pm$ 0.15 & 6.38 $\pm$ 0.63 \\
 & iMTFA~\cite{Ganea-CVPR2021} & 18.22 $\pm$ 0.41 & 22.56 $\pm$ 0.47 & 5.19 $\pm$ 0.44\\
 & iFS-RCNN~\cite{Nguyen-CVPR2022}  & 29.89 $\pm$ 0.09 & 36.33 $\pm$ 0.01 & 8.80 $\pm$ 0.50 \\
\rowcolor{lightgray} & \textbf{MaskDiff (Ours)} & 30.48 \textbf{{$\pm$ 0.06}} & 38.12 \textbf{{$\pm$ 0.0076}} & 9.43 \textbf{{$\pm$ 0.18}} \\
\midrule
\multirow{6}{*}{10} & MRCN+ft-full~\cite{He-ICCV2017} & 12.15 $\pm$ 0.33 & 18.08 $\pm$ 0.39 & 1.86 $\pm$ 0.27 \\
 & MTFA~\cite{Ganea-CVPR2021} & 25.97 $\pm$ 0.16 & 31.84 $\pm$ 0.25 & 8.36 $\pm$ 0.49\\
 & iMTFA~\cite{Ganea-CVPR2021} & 17.87 $\pm$ 0.28 & 21.87 $\pm$ 0.34 & 5.88 $\pm$ 0.45\\
 & iFS-RCNN~\cite{Nguyen-CVPR2022}  & 30.41 $\pm$ 0.06 & 36.32 $\pm$ 0.01 & 10.06 $\pm$ 0.37 \\
 \rowcolor{lightgray} & \textbf{MaskDiff (Ours)} & 31.89 \textbf{{$\pm$ 0.03}} & 38.55  \textbf{{$\pm$ 0.0062}} & 11.84 \textbf{{$\pm$ 0.15}} \\
\bottomrule
\end{tabular}
}
\caption{Stability of FSIS performance on COCO-All dataset. MaskDiff is the most stable method with smallest standard deviation of AP. The best performance is marked in boldface.}
\label{table:stability}
\end{table}

\begin{figure*}
     \centering
     \begin{subfigure}[b]{0.32\textwidth}
         \centering
         \includegraphics[width=\textwidth]{figures/IoU-1shot.pdf}
         \caption{$1$ shot}
     \end{subfigure}
     \hfill
     \begin{subfigure}[b]{0.32\textwidth}
         \centering
         \includegraphics[width=\textwidth]{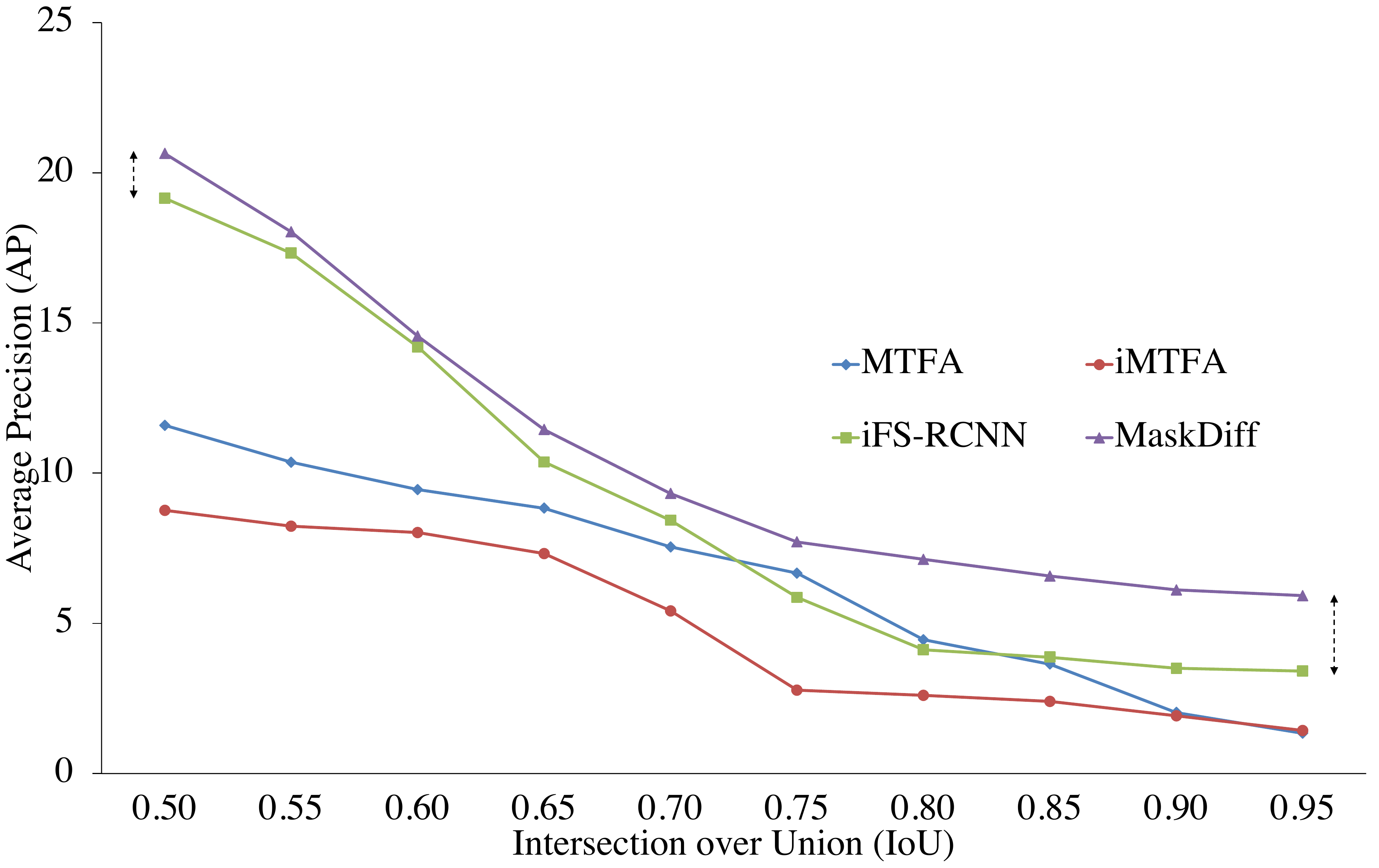}
         \caption{$5$ shots}
     \end{subfigure}
     \hfill
     \begin{subfigure}[b]{0.32\textwidth}
         \centering
         \includegraphics[width=\textwidth]{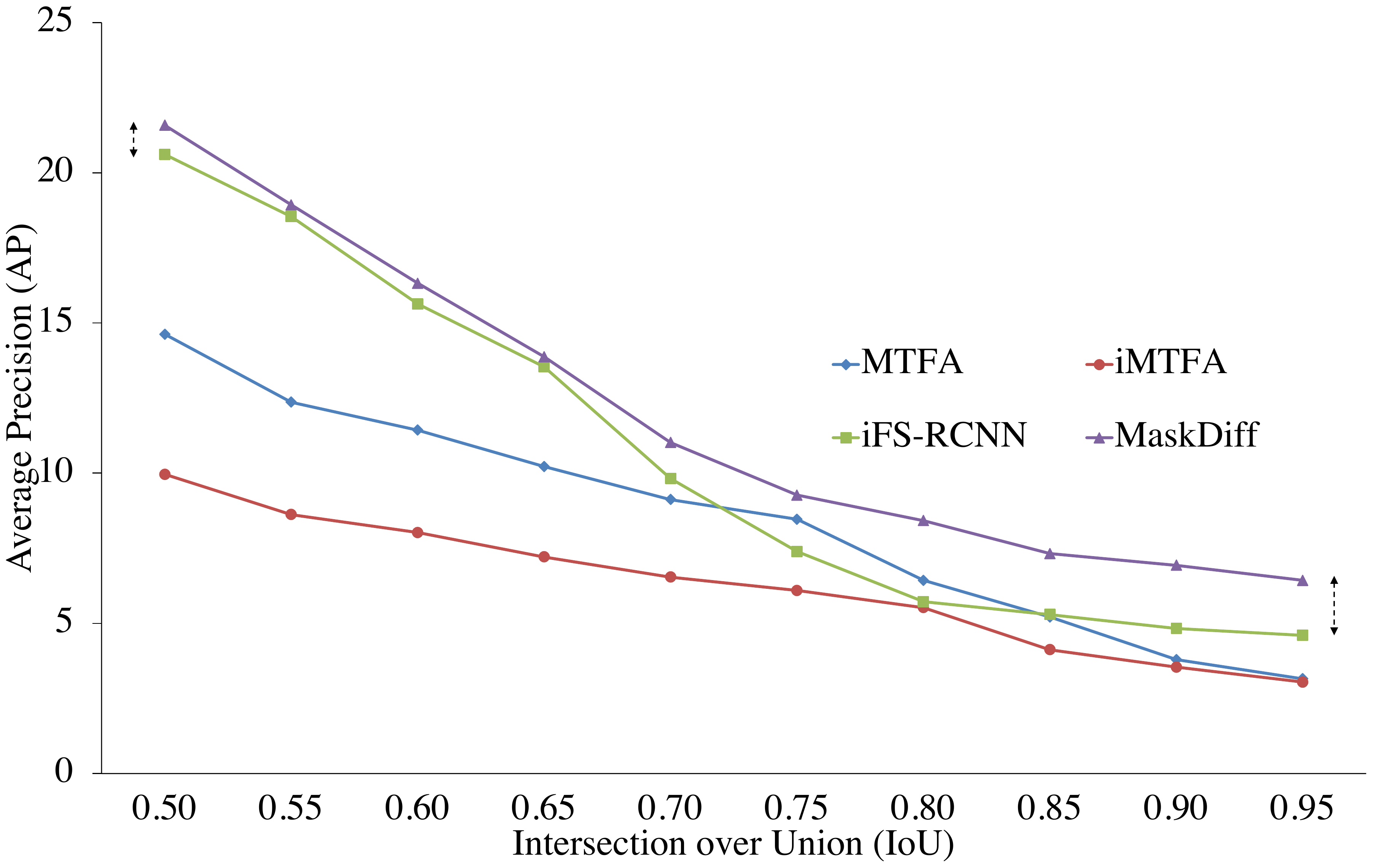}
         \caption{$10$ shots}
     \end{subfigure}
        \caption{MaskDiff preserves spatial information, especially at detailed levels. We compare the performance in terms of AP of MaskDiff with state-of-the-art methods in $K=1,5,10$ shots instance segmentation at different IoU thresholds. MaskDiff outperforms other methods with large margins, especially at high IoU thresholds, which indicates its ability to segment objects more precisely.}
        \label{fig:iou}
    \vspace{-3mm}
\end{figure*}

\textbf{Spatial information preservation.} Most conventional FSIS methods leverage pooled object features and feed them into Fully Convolutional Network (FCN)~\cite{Long-CVPR2015} to generate binary masks. However, passing through several pooling layers leads to a severe collapse in spatial information, especially the detailed one. By directly utilizing object regions from input images to generate binary masks, our MaskDiff can embrace comprehensive spatial details and induce precise delineation. In Fig.~\ref{fig:iou}, we visualize the line graph representing AP from $0.50$ to $0.95$ with step size $0.05$ to illustrate the changes in the performance of FSIS methods when the IoU requirement rises. Our MaskDiff surpasses cutting-edge methods, especially at high IoU thresholds.

\textbf{Effect of diffusion steps.} We do experiments on smaller diffusion steps to evaluate the effect of diffusion steps on the performance of our MaskDiff. Table~\ref{table:diffusion_steps} indicates that decreasing diffusion steps downgrades the effectiveness and stability of our method. However, MaskDiff with $250$ steps is still better than the SOTA method iFS-RCNN~\cite{Nguyen-CVPR2022}.
\begin{table}[!h]
\centering
\resizebox{0.8\columnwidth}{!}{
\begin{tabular}{l|c|c|c}
\midrule
 \textbf{Method} & \textbf{All Classes} & \textbf{Base Classes} & \textbf{Novel Classes} \\  
 \midrule
 iFS-RCNN~\cite{Nguyen-CVPR2022}  & 28.45 $\pm$ 0.120 & 36.35 $\pm$ 0.0110 & 3.95 $\pm$ 0.480\\
MaskDiff (250 steps) & 29.26 $\pm$ 0.094 & 37.18  $\pm$ 0.0095 & 4.57 $\pm$ 0.250 \\
 MaskDiff (500 steps) & 29.38 $\pm$ 0.093 & 37.55 $\pm$ 0.0095 & 4.68 $\pm$ 0.247 \\
\textbf{MaskDiff (1000 steps)} & \textbf{29.42} \textbf{{$\pm$ 0.091}} & \textbf{37.59} \textbf{{$\pm$ 0.0093}} & \textbf{4.85} \textbf{{$\pm$ 0.243}} \\
\midrule
\end{tabular}
}
\vspace{-3mm}
\caption{Stability and effectiveness of FSIS ($1-$shot) on COCO-All dataset with different diffusion steps.}
\label{table:diffusion_steps}
\vspace{-3mm}
\end{table}

\bibliography{aaai24}